\def\paperpreprintversion{1}

\documentclass{article} 
\usepackage{iclr2027_conference,times}

\usepackage{amsmath,amsfonts,bm}

\def\eqref#1{equation~\ref{#1}}

\def\1{\bm{1}}

\DeclareMathAlphabet{\mathsfit}{\encodingdefault}{\sfdefault}{m}{sl}
\SetMathAlphabet{\mathsfit}{bold}{\encodingdefault}{\sfdefault}{bx}{n}

\newcommand{\KL}{D_{\mathrm{KL}}}

\DeclareMathOperator*{\argmin}{arg\,min}

\usepackage{amsmath,amssymb,mathtools}
\usepackage{booktabs}
\usepackage{graphicx}
\usepackage{wrapfig}
\usepackage{placeins}
\usepackage{microtype}
\usepackage{url}
\usepackage{xcolor}
\usepackage{algorithm}
\usepackage{algpseudocode}
\usepackage{hyperref}
\usepackage{bbm}
\usepackage{enumitem}

\title{When Can Attention Heads Be Statically \\Defined?}

\author{\textbf{Weixian Waylon Li$^{1}$ \quad Yintao Tai$^{1}$ \quad Marcio Fonseca$^{2}$ \quad Shay B. Cohen$^{1}$} \\
{\normalfont $^{1}$University of Edinburgh, United Kingdom} \quad  $^{2}$\normalfont{Chamber of Deputies, Brazil} \\
\normalfont\fontfamily{lmtt}\selectfont
\{waylon.li,yintao.tai\}@ed.ac.uk \quad marcio.fonseca@camara.leg.br \quad \\
\normalfont\fontfamily{lmtt}\selectfont scohen@inf.ed.ac.uk
} 

\newcommand{\ourapproach}{\textsc{SAF}}
\newcommand{\fullourapproach}{Selective Attention Freezing}

\newcommand{\attn}{\mathbf{A}}
\newcommand{\meanattn}{\overline{\mathbf{A}}}
\newcommand{\prior}{\mathbf{P}}

\newcommand{\deltappl}{\Delta\mathrm{PPL}}
\definecolor{qualityWorseColor}{HTML}{A44A56}
\definecolor{qualityBetterColor}{HTML}{2F705D}
\newcommand{\qdegrade}[1]{\textcolor{qualityWorseColor}{#1}}
\newcommand{\qimprove}[1]{\textcolor{qualityBetterColor}{#1}}

\ifdefined\paperpreprintversion
    \iclrfinalcopy
\fi

\begin{document}

\maketitle
\ifdefined\paperpreprintversion
    \lhead{Preprint}
\fi

\begin{abstract}
Some attention heads learn similar patterns across inputs. Reusing these patterns could reduce training cost by avoiding repeated query-key score computation and softmax.
Through controlled pretraining comparisons, we identify \fullourapproach{} (\ourapproach{}), which selects heads with low attention-pattern variance and replaces their attention weights with fitted post-softmax means halfway through training.
We represent these fixed patterns with absolute-position and relative-distance preferences, reducing storage from quadratic to linear in sequence length.
A fused kernel reconstructs the patterns and executes ordinary-attention and replaced heads together.
At matched training-token budgets, replacing 25\% of attention heads gives 1.056$\times$ faster post-replacement optimiser updates at 124M parameters and 4K context, with a 0.77\% perplexity increase.
At 1B and 8K context, post-replacement updates are 1.068$\times$ faster on four GPUs including communication, with a 0.51\% perplexity increase.
The resulting models also accelerate long-input finetuning and causal prefill.
After associative-recall adaptation, the 124M model with 25\% replacement generalises to more key-value pairs at a fixed length better than ordinary attention and two pruning controls.
\end{abstract}

\section{Introduction}
\label{sec:introduction}

Self-attention computes its attention weights for every head and every input \citep{vaswani2017attention}. 
For each head, query-key interactions determine an input-dependent attention matrix, allowing the model to dynamically decide which tokens should interact. 
This input-dependence is the source of the mechanism's expressivity, but it comes at a computational cost: each head requires query and key projections, attention-score computation, and softmax normalisation, together with their backward passes during training, giving quadratic cost in sequence length.

For some heads, however, the attention weights depend primarily on query and key positions rather than on the tokens themselves, making their scores nearly content-free. 
Positional encodings make such heads easy to form, as shown for RoPE \citep{round2025}, and fixed positional patterns can replace learned heads with little loss in translation and pretrained encoders \citep{raganato2020fixed,tay2021synthesizer,hassid2022much}. 
Recent language-model studies also examine relaxed sequence dependence and random input-independent token mixing \citep{xue-etal-2025-deconstructing,dong2025random}.
For a head whose scores no longer depend on content, computing them for every input reproduces the same positional bias while still paying for the computational cost.

\begin{figure}
    \centering
    \includegraphics[width=1\linewidth]{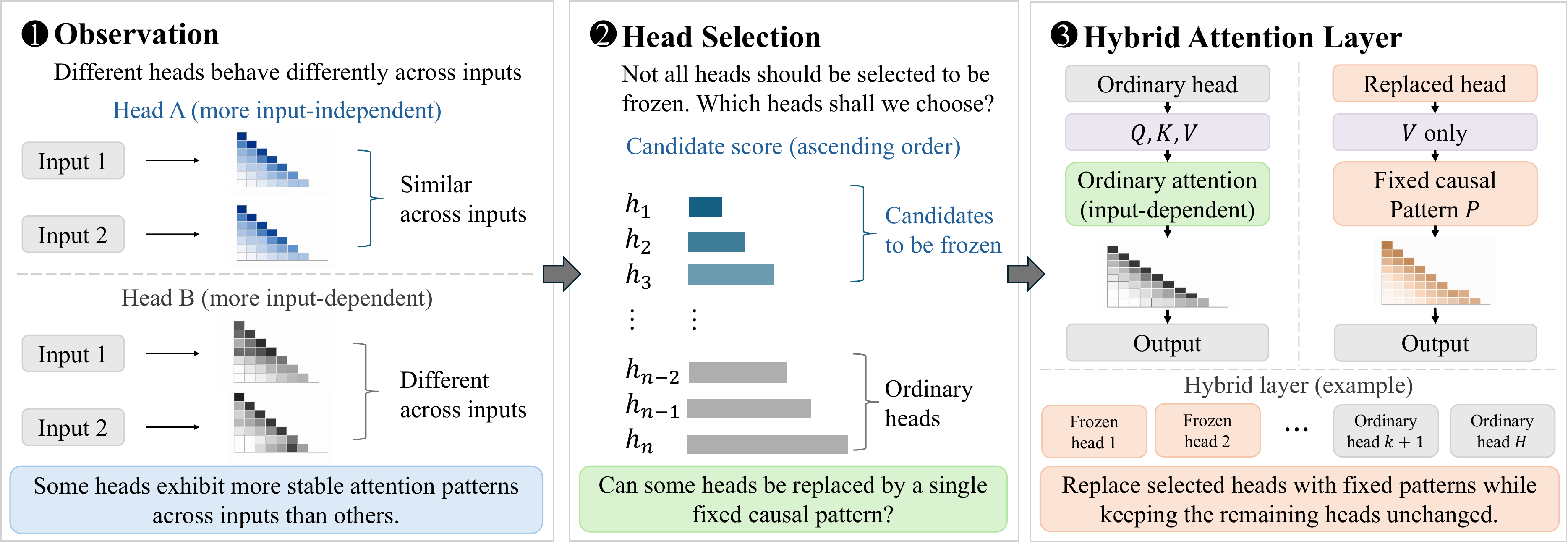}
    \caption{Overview of fixed-pattern replacement of ordinary attention.}
    \label{fig:overview}
    \vspace{-0.5cm}
\end{figure}

This observation means we can replace such heads during training, but existing results remains unclear which heads and patterns to choose, when to replace them, and if a layer that mixes fixed and ordinary-attention heads yields memory and wall-clock savings with modern kernels.
So we ask:

\begin{center}
    \emph{(1) Under what conditions can an input-dependent attention matrix be replaced by a fixed pattern and (2) how does this improve computational and memory efficiency?}
\end{center}
Our controlled comparisons lead to \fullourapproach{} (\ourapproach{}), a recipe that replaces selected attention matrices with fixed causal patterns.
\ourapproach{} fixes the attention weights but retains token mixing of input-dependent values: the value ($\mathbf{V}$) and output projections remain trainable.
The hybrid layer combines ordinary query-key attention with fixed-pattern heads (Figure~\ref{fig:overview}).
We compare patterns, selectors, rates, and timing during pretraining, including pruning at matched tokens and time, then evaluate finetuning and causal prefill.

A dense fixed attention matrix requires quadratic space.
Because a content-free pattern is a positional bias, we store it as absolute-key-position and relative-distance preferences in vectors of linear size, and develop a fused kernel that executes ordinary attention with FlashAttention \citep{dao2022flashattention} and frozen heads in one forward pass.
Reconstructing fixed patterns in registers bypasses softmax, and $\mathbf{Q}$/$\mathbf{K}$ gradients in our multi-head pretraining path, reducing pattern bandwidth and activation storage.

At 25\% replacement, \ourapproach{} increases perplexity by $0.77\pm0.07\%$ at 124M and 4K context, with $1.056\times$ faster post-replacement updates.
At 1B and 8K, the increase is $0.51\%$ with a $1.068\times$ speedup on four GPUs, including communication.
These savings accumulate over the remaining training updates, reducing the accelerator time needed to process a fixed token budget.
The 50\% setting offers larger speed and memory gains with a higher perplexity cost.
Longer-context 124M models have smaller perplexity penalties and accelerate long-input finetuning and causal prefill.
After adaptation on eight key-value pairs, the 124M \ourapproach{} models outperform both pruning baselines on 24--64 pairs.

\section{Related Work}
\label{sec:related-work}

\paragraph{Input-independent and constrained attention.}
Fixed positional patterns in translation \citep{raganato2020fixed,you-etal-2020-hard}, input-independent audio encoders \citep{wu2020inputindependent}, and learned or random synthetic token mixing \citep{tay2021synthesizer,dong2025random} show that useful computation can persist without standard attention.
PAPA uses input-averaged attention in pretrained encoders \citep{hassid2022much}, and recent work relaxes sequence dependence in language models \citep{xue-etal-2025-deconstructing}. Theory also studies expressivity under frozen weights or sparse graphs \citep{zaheer2020bigbird,fu2023single,otsuka2025strong}.
We build on input-averaged attention through selective replacement during causal language-model training, compact storage, and fused execution.

\paragraph{Head heterogeneity and efficient training.}
Distinct head functions motivate pruning and component-importance measures \citep{NEURIPS2019_2c601ad9,voita-etal-2019-analyzing,men2024shortgpt}.
FLAP uses activation fluctuations to guide structured pruning and compensates removed features with a constant bias \citep{An_2024}.
Fixed-pattern replacement instead retains token mixing of the current input's values.
Retrieval-head analysis studies key-value recall \citep{wu2025retrieval}; DuoAttention separates retrieval and streaming heads \citep{xiao2025duoattention}, while MInference assigns head-specific sparse patterns \citep{jiang2024minference}.
Related training strategies freeze or prune components progressively \citep{brock2017freezeout,zhang2020progressive,liu2021autofreeze,xia2024sheared,erdogan2025layerlock}, and quantisation-aware training also exposes compute-allocation trade-offs \citep{dremov2026computeoptimal}.

\paragraph{Hardware-aware execution.}
FlashAttention provides efficient exact attention \citep{dao2022flashattention,dao2023flashattention2}, while MInference and DuoAttention exploit head heterogeneity within kernels \citep{jiang2024minference,xiao2025duoattention}.
Our kernel executes ordinary attention and fixed-pattern token mixing together, removing score computation and softmax for replaced heads.
The pretraining path also omits their query and key projections; projection savings depend on whether keys are shared (Section~\ref{sec:fused-kernel}).
\section{Replacing Attention with a Fixed Pattern}
\label{sec:method}
\label{sec:intervention}

After an initial period of training with ordinary attention, we use a set of calibration inputs to estimate attention statistics at the current checkpoint.
We then select heads, construct fixed patterns, and resume training.
Algorithm~\ref{alg:calibration} gives the resulting one-time \ourapproach{} procedure.

Let $L$ denote the number of attention layers and $H$ the number of query heads per layer, giving $LH$ heads in total. We use $T$ for sequence length and $d_h$ for head dimension.
For normalised layer input $\mathbf{X}_{\ell}$, let $\mathbf{Q}_{\ell h},\mathbf{K}_{\ell h},\mathbf{V}_{\ell h}$ be the head projections and $\mathbf{S}_{\ell h}=\mathbf{Q}_{\ell h}\mathbf{K}_{\ell h}^{\top}/\sqrt{d_h}$ the attention scores.
Ordinary attention computes input-dependent query--key scores and mixes values using $\attn_{\ell h}(\mathbf{X}_{\ell})=\operatorname{softmax}_{\mathrm{causal}}(\mathbf{S}_{\ell h})$.
With $i,j$ indexing query and key positions, we replace a selected head's attention weights $\attn_{\ell h}$ by a fixed causal row-stochastic pattern $\prior_{\ell h}\in\mathbb{R}^{T\times T}$, giving the head output: $ \widetilde{\mathbf{Z}}_{\ell h}(\mathbf{X}_{\ell}) = \prior_{\ell h}\mathbf{V}_{\ell h}$.

Both paths perform token mixing: a weighted sum of value vectors across token positions.
Value and output projections remain trainable, so the head output still depends on the input through $\mathbf{V}_{\ell h}$; only its attention weights are fixed.
For replaced heads in multi-head attention, query and key projections, score computation, and softmax are removed.
Appendix~\ref{app:attention-notation} gives full tensor definitions.

\subsection{Calibration and head selection}
\label{sec:calibration-selection}

Attention heads exhibit different patterns and roles \citep{NEURIPS2019_2c601ad9,voita-etal-2019-analyzing,clark-etal-2019-bert,li-etal-2023-bert,neo-etal-2024-interpreting,jiang2024minference,li2026spectral}.
We compare scores based on attention variation and head outputs.
At the replacement checkpoint, we measure attention on $N$ calibration sequences in evaluation mode, leaving model weights and the training-data sampler unchanged.
For calibration input $\mathbf{X}_{\ell,n}$, write $\attn_{\ell h}^{(n)}=\attn_{\ell h}(\mathbf{X}_{\ell,n})$ and $\mathbf{S}_{\ell h}^{(n)}=\mathbf{S}_{\ell h}(\mathbf{X}_{\ell,n})$; the empirical mean is $\widehat{\attn}_{\ell h}=N^{-1}\sum_n\attn_{\ell h}^{(n)}$.
We rank heads globally by increasing score $s_{\ell h}$, targeting $k=\operatorname{round}(rLH)$ heads at replacement rate $r$.
The attention-variance score is
\begin{equation}
    s_{\ell h}^{\mathrm{var}}=\frac{1}{(N-1)T^2}\sum_{n=1}^{N}
    \left\lVert\attn_{\ell h}^{(n)}-\widehat{\attn}_{\ell h}\right\rVert_F^2,
    \label{eq:variance-score}
\end{equation}
where $\lVert\cdot\rVert_F$ is the Frobenius norm.
Calibration data are disjoint from reporting data (Appendix~\ref{app:attention-notation}).
The variance score measures reconstruction error across inputs, rather than functional importance, and also depends on attention concentration.
The forward-KL score instead compares each causal query-row distribution with its empirical mean:
\begin{equation}
    s_{\ell h}^{\mathrm{KL}}
    =\frac{1}{NT}\sum_{n=1}^{N}\sum_{i=1}^{T}
    \mathrm{KL}\!\left(\attn_{\ell h}^{(n)}[i,1{:}i]\,\middle\|\,
    \widehat{\attn}_{\ell h}[i,1{:}i]\right).
    \label{eq:forward-kl-score}
\end{equation}
It measures input dependence in probability space, accumulating directly from row entropies during the same calibration pass. 
Appendix~\ref{app:selector-schedule-definitions} introduces alternative selection metrics, including Q/K-gradient scores, output magnitude, within-layer redundancy, and residual influence. 
Appendix~\ref{app:head-selection-analysis} evaluates these alternatives by measuring the immediate loss penalty of single-head replacements and the long-term model perplexity after continued training.

For compact patterns (Section~\ref{sec:compact-prior}), we consider candidates in score order and skip fits whose row-averaged KL error exceeds $\varepsilon_{\mathrm{fit}}$.
Selection continues until $k$ fits pass or no candidates remain. 
The score determines candidate order; $\varepsilon_{\mathrm{fit}}$ bounds compact-fitting error.

\subsection{Fixed-pattern construction and fitting}
\label{sec:candidate-choices}
\label{sec:compact-prior}

For each candidate head, we compare five causal row-stochastic patterns: deterministic averages, fitted distributions, and a structured-random control.
\begin{itemize}[topsep=0pt,leftmargin=*]
    \item \textbf{Post-softmax mean.} $\prior_{\ell h}=\widehat{\attn}_{\ell h}$ averages attention probabilities across inputs and remains causal and normalised.
    \item \textbf{Sharp mean.} $\prior_{\ell h}=\operatorname{softmax}_{\mathrm{causal}}(N^{-1}\sum_n\mathbf{S}_{\ell h}^{(n)})$ averages logits before softmax.
    \item \textbf{Gaussian sample.} We sample causal logits independently from normal distributions fitted to their means and variances across inputs, then apply causal softmax. This captures marginal logit variation but discards correlations.
    \item \textbf{Dirichlet sample.} We sample each causal row jointly with mean $\widehat{\attn}_{\ell h}[i,1{:}i]$ and concentration fitted from across-input variances. The draws are nonnegative and normalised without softmax.
    \item \textbf{Structured random (control).} We sample $\alpha_{\ell h}(j),\rho_{\ell h}(\delta)\overset{\mathrm{iid}}{\sim}\mathcal{N}(0,1)$ and construct Equation~\ref{eq:compact-prior}, testing random preferences within the same positional parameterisation.
\end{itemize}
Each sampled pattern is drawn once at replacement and retained thereafter. Appendix~\ref{app:dirichlet-fit} gives distribution-fitting details and empirical checks.

\paragraph{Compact representation.}
Dense patterns require $O(T^2)$ storage per head.
We represent absolute-key-position and relative-distance preferences with length-$T$ vectors $\alpha$ and $\rho$, motivated by distance structure and attention sinks \citep{sink2024,round2025,yang2026patterns}.
Suppressing head indices, $\alpha$ uses key indices $1,\ldots,T$ and $\rho$ uses offsets $0,\ldots,T-1$.
With $Z(i)=\sum_{k=1}^{i}\exp(\alpha(k)+\rho(i-k))$, the pattern is
\begin{equation}
    \widehat{\prior}(i,j)
    =\begin{cases}
        \displaystyle\frac{\exp(\alpha(j)+\rho(i-j))}{Z(i)}, & j\leq i, \\[4pt]
        0, & j>i.
    \end{cases}
    \label{eq:compact-prior}
\end{equation}
Each query row is normalised separately. The vectors $\alpha$ and $\rho$ capture absolute-position effects such as attention sinks and query-key distance, respectively, using $O(T)$ storage per head.

\paragraph{Fitting.}
We fit the compact pattern in Equation~\ref{eq:compact-prior} to a target fixed pattern $\prior$ by minimising row-averaged cross-entropy:
\begin{equation}
    \mathcal{L}_{\mathrm{fit}}(\alpha,\rho)
    =\frac{1}{T}\Big[
      \sum_i\log Z(i)-\sum_j\eta_{\mathrm{abs}}(j)\,\alpha(j)
      -\sum_{\delta}\eta_{\mathrm{rel}}(\delta)\,\rho(\delta)
    \Big],
    \label{eq:fit-objective}
\end{equation}
where $\eta_{\mathrm{abs}}(j)=\sum_{i\geq j}\prior(i,j)$ and $\eta_{\mathrm{rel}}(\delta)=\sum_{i-j=\delta}\prior(i,j)$ are the target's absolute-position and relative-distance marginals.
These two length-$T$ marginals are sufficient statistics for fitting. Their linearity permits online accumulation during calibration, although our implementation first collects dense attention statistics.
We use a fitting tolerance of $\varepsilon_{\mathrm{fit}}=0.2$ nats per row. Comparisons with prescribed head sets require every fit to pass, without substituting other heads (Appendix~\ref{app:training-algorithm}).
Appendix~\ref{app:kernel-details} gives the optimality conditions and memory-bounded fitter.

\subsection{Replacement schedules and fused execution}
\label{sec:fused-kernel}

\begin{wrapfigure}{R}{0.45\textwidth}
    \vspace{-0.5cm}
    \centering
    \includegraphics[width=0.98\linewidth]{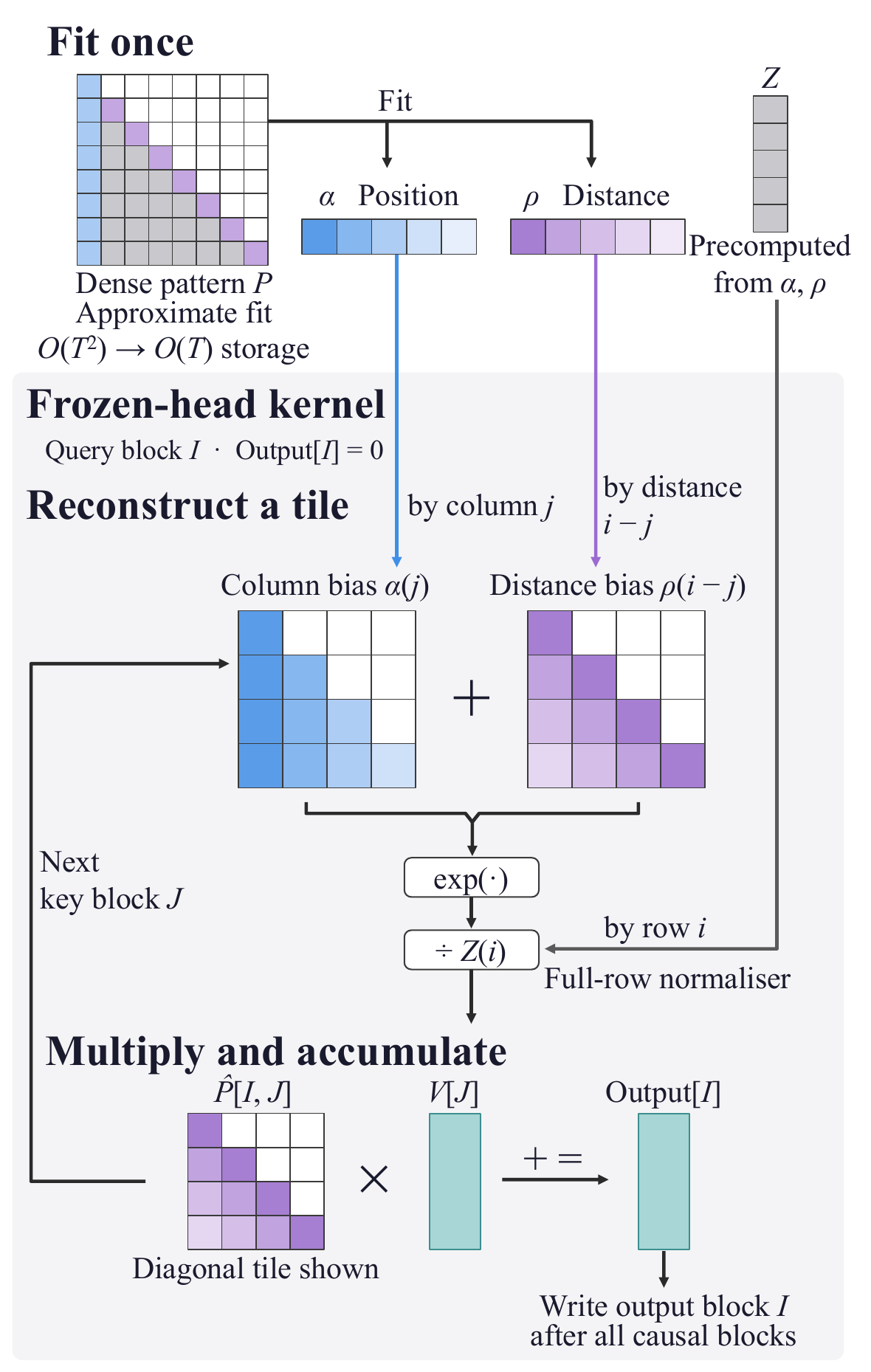}
    \caption{Compact representation and execution of a fixed attention pattern.}
    \label{fig:compact-kernel}
    \vspace{-0.5cm}
\end{wrapfigure}

We store the fixed patterns as buffers and resume training from the same model weights and optimiser state.
Ordinary-attention heads and the remaining parameters, including value and output projections, continue to train.
$\alpha$, $\rho$, and the normalisers receive no gradients.

\paragraph{Replacement schedules.}
Gradual and iterative pruning strategies motivate varying both the extent and timing of replacement \citep{xia2024sheared,shen-etal-2024-pruning,wang-etal-2026-local}.
One-time replacement freezes the entire selected set at one update. Gradual replacement installs the same final heads in nearly equal groups at evenly spaced updates, testing whether smaller interventions ease adaptation.
The validation-budget schedule controls cumulative measured validation-perplexity cost, while the train-loss-triggered schedule uses training-loss plateaux to propose replacements and waits for recovery between groups. Appendix~\ref{app:selector-schedule-definitions} formalises these schedules and details their settings.

\paragraph{Fused computation.}
Our fused kernel executes ordinary-attention and replaced heads in one forward launch per layer.
A head-state flag selects each kernel program's computation, avoiding separate launches and concatenation.
Ordinary-attention heads use FlashAttention-2's online-softmax tiling \citep{dao2022flashattention,dao2023flashattention2}.
Replaced heads reconstruct causal tiles of $\widehat{\prior}$ in registers from $\alpha$, $\rho$, and precomputed normalisers, then multiply them with value tiles (Figure~\ref{fig:compact-kernel}).
This avoids score computation, softmax, and dense pattern reads for replaced heads.
Their attention backward pass is $\mathrm{d}\mathbf{V}=\widehat{\prior}^{\top}\mathrm{d}\mathbf{Z}$.
Our multi-head pretraining implementation omits replaced query and key projections; the Qwen grouped-query path omits replaced query projections but retains full shared key and value projections.
Storage is $O(T)$ per replaced head, while token mixing remains $O(T^2d_h)$.
Appendix~\ref{app:kernel-details} gives execution details and numerical validation; Section~\ref{sec:across-stages} measure update and prefill costs.

\section{Experimental Setup}
\label{sec:experimental-setup}

\paragraph{Model pretraining.}
We train a 124M decoder with 12 Transformer layers, 12 heads per layer, and hidden dimension 768 on FineWeb-Edu \citep{penedo2024fineweb}, comparing learned absolute positions with RoPE \citep{su2023roformer} at 4K context.
Each run performs 5,000 updates of 491,520 target tokens, totalling 2.4576B tokens or approximately 20 tokens per parameter \citep{hoffmann2022chinchilla}. 
Replacement runs share the source checkpoint, training configuration, and subsequent data order. Pattern comparisons use identical head sets; rate comparisons use nested subsets of one ranking. Replacement occurs at the midpoint unless timing is varied.
Gaussian, Dirichlet, and sharp patterns use dense storage in the pattern comparison. The post-softmax mean is tested in dense and compact forms; systems measurements use the compact representation.
At 4K, calibration uses 32 sequences from the training corpus, sampled with a separate random-number generator so that the training-data order is unchanged. Calibration inputs are disjoint from final reporting data (Appendix~\ref{app:attention-notation}).

\paragraph{Control baselines.} 
Three-seed controls compare pruning and random head selection with mean replacement at matched tokens.
We use micro-batch size 8, 15 accumulation steps, and shuffled non-overlapping training windows. 
Random selection retains fitted means and matches per-layer head counts. 
These controls use 32 held-out calibration sequences (Appendix~\ref{app:fixed-token-controls}).
At matched time, continuations share an allowance and elapsed-time learning-rate schedule, so faster models process more tokens.
We also compare frozen and trainable compact mean patterns from identical fitted initialisations, and evaluate both in multi-query associative recall (MQAR; Appendix~\ref{app:pattern-controls}; \citealp{ICLR2024_448fc91f}).
Gate-Taylor pruning scores each head by the mean absolute loss gradient with respect to a scalar gate on its output \citep{NEURIPS2019_2c601ad9}. 
We normalise scores within each layer and remove the globally lowest-scoring heads once (Appendix~\ref{app:pruning-selection-budget}).

\paragraph{Quality and runtime.}
For replaced and ordinary-attention held-out perplexities $p_{\mathrm{rep}}$ and $p_{\mathrm{ord}}$, we report $\deltappl=100(p_{\mathrm{rep}}/p_{\mathrm{ord}}-1)$.
We compare ordinary-attention and replaced checkpoints on SST-2 \citep{socher-etal-2013-recursive}, BoolQ \citep{clark-etal-2019-boolq}, and QuALITY \citep{pang-etal-2022-quality} over three finetuning seeds each.
Pruning and MQAR \citep{ICLR2024_448fc91f} comparisons vary pretraining and task-training seeds separately (Appendices~\ref{app:pruning-selection-budget} and~\ref{app:associative-recall}).
All systems measurements use GH200 GPUs with bfloat16 computation.
Pretraining update timings include gradient accumulation, backward computation, clipping, AdamW, and gradient reset, but exclude data loading, calibration, scoring, and fitting.
The supplementary controls report total training times including the initial training with ordinary attention and intervention.
Speedups aggregate within-pair ratios of ordinary-attention to replaced-model update times; displayed times are aggregated separately, using medians for single-GPU pretraining benchmarks and means for finetuning and four-GPU benchmarks.
Quality and runtime use separate runs in the 124M study (Appendices~\ref{app:controlled-pretraining-details} and~\ref{app:downstream-details}).

\paragraph{Transfer evaluations.}
The 8K and 16K runs retain the 2.4576B-token budget, training models and fitting patterns separately at each length. Their checkpoints provide square causal-prefill benchmarks across lengths and batch sizes, excluding token-by-token KV-cache decoding.
Qwen3-4B is evaluated on five zero-shot tasks without further training; variance and forward-KL selection each use 512 calibration sequences (Appendix~\ref{app:pretrained-static-details}).
At 1B parameters and 8K context, ordinary-attention, replaced, and Gate-Taylor-pruned models are applied on the same midpoint checkpoint, and a 19.667B-token budget on four GH200 GPUs.
We compare matched held-out PPL and three-seed finetuning, and benchmark ordinary-attention and replaced updates with four-GPU communication (Appendix~\ref{app:scale-up-details}).

\section{Results}
\label{sec:results}

\subsection{Which fixed patterns and heads should be used?}
\label{sec:pattern-results}

\begin{wrapfigure}{R}{0.47\textwidth}
    \vspace{-0.8\baselineskip}
    \centering
    \includegraphics[width=0.98\linewidth]{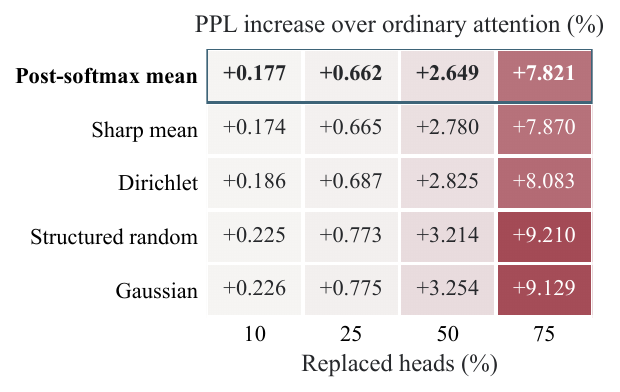}
    \vspace{-0.2cm}
    \caption{Perplexity increases for five fixed patterns and four replacement rates.}
    \label{fig:prior-quality}
     
\end{wrapfigure}

\paragraph{Fixed-pattern content.}
At matched tokens, the post-softmax mean gives the lowest or similar perplexity across replacement rates, followed most closely by the sharp mean (Figure~\ref{fig:prior-quality}).
Differences are small up to 25\% replacement but widen at higher rates; neither Gaussian nor Dirichlet sampling improves on the mean.
Appendices~\ref{app:dirichlet-fit} and~\ref{app:recipe-analysis} give diagnostics and reconstruction analysis.

\paragraph{Head selection.}
Variance selection gives lower perplexity after continued training than residual cosine, although residual cosine better predicts the immediate cost of replacing one head (Appendix~\ref{app:head-selection-analysis}).
Projected-output magnitude and redundancy are weaker predictors of this local cost (Table~\ref{tab:selector-diagnostic}).
Forward KL gives a larger perplexity penalty after adaptation at every tested context and rate (Table~\ref{tab:kl-continuation-selector}).
We therefore retain attention variance for replacement during pretraining.
At 25\% replacement, selected heads concentrate in earlier layers and have more diffuse attention than unselected heads in those layers. 
Across three seeds, 30--31 of the 36 heads selected at quarter-training remain selected at the midpoint (Appendix~\ref{app:head-selection-analysis}).
With layer distributions matched, selected heads have higher mean normalised entropy (0.837 versus 0.678) and place less attention on the most recent 64 tokens (19.0\% versus 40.7\%).
Low variance therefore favours diffuse attention, but its association with immediate replacement cost remains positive after controlling for both layer and entropy (partial Spearman correlation 0.543).
Appendix~\ref{app:head-selection-analysis} gives the profile and one-head diagnostic protocols.

\paragraph{Head selection and retained token mixing.}
Random selection nearly doubles the 25\% perplexity penalty despite matching per-layer head counts (Table~\ref{tab:fixed-token-controls-main}).
Variance selection also lowers the matched-time penalty from 1.24\% to 0.46\%, showing that its benefit extends beyond allocating more fixed heads to earlier layers.

Continuing to optimise fitted $\alpha,\rho$ reduces perplexity by only $0.010\pm0.006\%$ relative to freezing, while increasing update time by $11.9\pm1.2\%$ at 25\% replacement (Table~\ref{tab:pattern-controls}a).
Keeping the fitted patterns fixed therefore gives nearly the same LLM performance with cheaper updates.

Retaining token mixing with the mean pattern gives lower perplexity than pruning the same low-variance heads at matched tokens.
At matched time, pruning completes more updates and has lower perplexity than replacement.
After longer training with ordinary attention, retaining token mixing still improves on same-head pruning across three seeds, but Gate-Taylor pruning gives lower perplexity under both budgets (Appendix~\ref{app:pruning-selection-budget}).
Low variance therefore identifies heads whose weights can be fixed, rather than a general ranking of heads to remove.

\begin{table}[t]
    \centering
    \footnotesize
    \setlength{\tabcolsep}{3pt}
    \caption{Perplexity increase (\%, lower is better) at matched tokens or time: mean $\pm$ SE over three seeds.
    Times include the initial training with ordinary attention and intervention.}
    \label{tab:fixed-token-controls-main}
    \begin{tabular}{@{}lrrrrrr@{}}
        \toprule
        & \multicolumn{4}{c}{Matched tokens: 2.4576B} & \multicolumn{2}{c}{Matched time} \\
        \cmidrule(lr){2-5}\cmidrule(l){6-7}
        & \multicolumn{2}{c}{$\deltappl$ (\%)} & \multicolumn{2}{c}{Time (min)} & \multicolumn{2}{c}{$\deltappl$ (\%)} \\
        \cmidrule(lr){2-3}\cmidrule(lr){4-5}\cmidrule(l){6-7}
        Operation + selection & 25\% & 50\% & 25\% & 50\% & 25\% & 50\% \\
        \midrule
        Mean + variance & $0.663\pm0.018$ & $2.495\pm0.024$ & 184.7 & 178.7 & $0.456\pm0.108$ & $1.911\pm0.112$ \\
        Pruning + variance & $0.806\pm0.018$ & $3.033\pm0.056$ & 178.9 & 168.3 & $0.113\pm0.169$ & $0.978\pm0.108$ \\
        Mean + random heads & $1.282\pm0.056$ & $3.630\pm0.093$ & 183.4 & 178.8 & $1.243\pm0.112$ & $2.778\pm0.125$ \\
        \bottomrule
    \end{tabular}
    \par\smallskip
    \raggedright Ordinary-attention PPL: 23.896 at matched tokens (188.7 min); 23.882 at matched time (187.65 min).
    Appendix~\ref{app:fixed-token-controls} gives absolute PPL, update counts, and paired contrasts.
     \vspace{-0.3cm}
\end{table}

\subsection{How many heads should be replaced, and when?}
\label{sec:rate-schedule-results}

\paragraph{Replacement rate.}
Replacement rate has the largest effect on model quality.
Across three pretraining seeds, midpoint replacement by the post-softmax mean increases perplexity by $0.768\pm0.067\%$ at 25\% and $2.492\pm0.083\%$ at 50\% (mean $\pm$ standard error; Table~\ref{tab:main-results}a).
We therefore prefer 25\% for retaining quality and use 50\% to examine the trade-off at a higher rate.

\paragraph{Replacement time.}

Replacement is robust across a broad range of training stages, with the largest penalties when little or no training remains (Table~\ref{tab:timing-main} in Appendix~\ref{app:controlled-pretraining-details}).
The penalty is 0.662--0.724\% for replacement between 25\% and 60\% of training, rising to 0.936\% at 75\% and 8.168\% after training. Continued training enables adaptation, with little sensitivity to the exact update across earlier stages.
Across three seeds, moving 25\% replacement from the midpoint to quarter-training raises the perplexity penalty from 0.663\% to 0.772\% and saves 100 seconds on average (Appendix~\ref{app:fixed-token-controls}).

The post-softmax mean outperforms structured-random patterns at every tested replacement time and rate, retaining its advantage after adaptation.

\paragraph{Fixed and automatic schedules.}
Neither gradual nor automatic replacement improves the quality--speed trade-off over one-time replacement (Tables~\ref{tab:schedule-details} and~\ref{tab:schedule-confirmation} in Appendix~\ref{app:controlled-pretraining-details}).
At matched 25\% and 50\% rates, the train-loss-triggered schedule achieves similar quality across three seeds but takes longer because it repeatedly measures attention.
Validation-loss budgets provide direct quality control but select substantially fewer heads under the tested budgets.
We use one-time midpoint replacement: these fixed-token comparisons show low perplexity penalties with a single measurement.

\paragraph{The \ourapproach{} recipe.}
These comparisons leads to \ourapproach{}: variance selection, compact mean patterns, and one-time midpoint replacement, with 25\% frozen rate.
The mean minimises fixed-matrix reconstruction error, and variance measures this minimum for each head (Appendix~\ref{app:recipe-analysis}), motivating both choices. 
Compact fitting changes final perplexity by less than 0.014\% relative to the dense mean under both position encodings; Appendix~\ref{app:kernel-details} reports agreement with the numerical reference.

\subsection{Does the trade-off carry across model use stages?}
\label{sec:across-stages}

\begin{table*}[t]
    \centering
    \scriptsize
    \setlength{\tabcolsep}{3.1pt}
    \caption{Quality and training costs for ordinary-attention and \ourapproach{} 124M checkpoints.
    Ord./repl. (ms) reports optimiser-update times for the ordinary-attention and replaced models.}
    \label{tab:main-results}
    \begin{tabular*}{\textwidth}{@{\extracolsep{\fill}}lcrrrrrr@{}}
        \toprule
        \multicolumn{2}{@{}l}{\emph{(a) Pretraining on FineWeb-Edu}} & \multicolumn{2}{c}{Perplexity $\downarrow$} & & \multicolumn{2}{c}{Update time} & Memory \\
        \cmidrule(lr){3-4}\cmidrule(lr){6-7}\cmidrule(l){8-8}
        Context ($B$) & Replaced heads & Ordinary attn. & Replaced & $\Delta$PPL (\%) & Ord./repl. (ms) & Speedup & Peak change \\
        \midrule
        4K (8)  & 25\% & 23.862 & 24.045 & \qdegrade{$+0.768\pm0.067\%$} & 2249.49/2129.42 & \qimprove{1.056$\times$} & \qimprove{$-1.96\%$} \\
        4K (8)  & 50\% & 23.862 & 24.457 & \qdegrade{$+2.492\pm0.083\%$} & 2229.80/1991.85 & \qimprove{1.119$\times$} & \qimprove{$-3.04\%$} \\
        8K (4)  & 25\% & 24.179 & 24.318 & \qdegrade{$+0.578\pm0.011\%$} & 2710.24/2636.22 & \qimprove{1.030$\times$} & \qimprove{$-1.63\%$} \\
        8K (4)  & 50\% & 24.179 & 24.676 & \qdegrade{$+2.058\pm0.066\%$} & 2782.97/2524.39 & \qimprove{1.101$\times$} & \qimprove{$-2.55\%$} \\
        16K (2) & 25\% & 24.428 & 24.538 & \qdegrade{$+0.450\pm0.012\%$} & 3741.42/3649.78 & \qimprove{1.025$\times$} & \qimprove{$-0.82\%$} \\
        16K (2) & 50\% & 24.428 & 24.865 & \qdegrade{$+1.792\pm0.112\%$} & 3697.14/3454.31 & \qimprove{1.070$\times$} & \qimprove{$-1.57\%$} \\
        \bottomrule
    \end{tabular*}
    \par\smallskip
    \begin{tabular*}{\textwidth}{@{\extracolsep{\fill}}lcrrrrrr@{}}
        \toprule
        \multicolumn{2}{@{}l}{\emph{(b) Supervised finetuning}} & \multicolumn{2}{c}{Accuracy (\%) $\uparrow$} & & \multicolumn{2}{c}{Update time} & Memory \\
        \cmidrule(lr){3-4}\cmidrule(lr){6-7}\cmidrule(l){8-8}
        Task ($B$) & Replaced heads & Ordinary attn. & Replaced & $\Delta$accuracy (pp) & Ord./repl. (ms) & Speedup & Peak change \\
        \midrule
        SST-2 (256) & 25\% & 86.39 & 86.54 & \qimprove{$+0.15\pm0.53\,\mathrm{pp}$} & 50.11/58.22 & \qdegrade{0.901$\times$} & \qimprove{$-3.65\%$} \\
        SST-2 (256) & 50\% & 86.39 & 86.77 & \qimprove{$+0.38\pm1.00\,\mathrm{pp}$} & 50.11/56.70 & \qdegrade{0.925$\times$} & \qimprove{$-5.95\%$} \\
        BoolQ (64) & 25\% & 71.40 & 70.62 & \qdegrade{$-0.77\pm0.56\,\mathrm{pp}$} & 67.36/70.91 & \qdegrade{0.967$\times$} & \qimprove{$-4.63\%$} \\
        BoolQ (64) & 50\% & 71.40 & 70.84 & \qdegrade{$-0.56\pm0.70\,\mathrm{pp}$} & 67.36/69.35 & \qdegrade{0.992$\times$} & \qimprove{$-7.19\%$} \\
        QuALITY 16K (4) & 25\% & 26.59 & 26.09 & \qdegrade{$-0.50\pm0.63\,\mathrm{pp}$} & 578.72/552.90 & \qimprove{1.047$\times$} & \qimprove{$-2.12\%$} \\
        QuALITY 16K (4) & 50\% & 26.59 & 26.94 & \qimprove{$+0.35\pm0.18\,\mathrm{pp}$} & 578.72/517.88 & \qimprove{1.118$\times$} & \qimprove{$-3.92\%$} \\
        \bottomrule
    \end{tabular*}
    \vspace{-0.4cm}
\end{table*}

\begin{figure*}[t]
    \centering
    \includegraphics[width=0.95\textwidth]{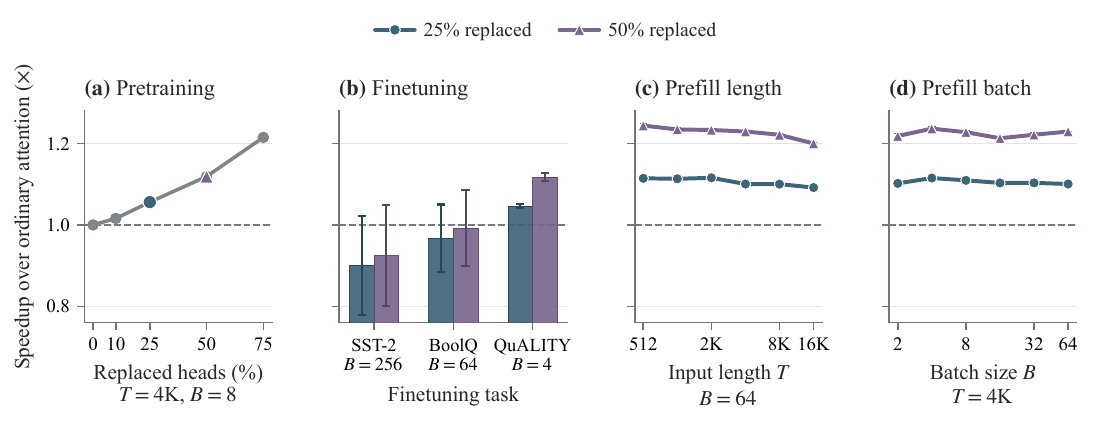}
    \caption{\ourapproach{} speedup over FlashAttention: (a) pretraining updates; (b) finetuning updates; (c) causal prefill by input length at $B=64$; (d) prefill by batch size at $T=4096$.}
    \label{fig:systems-across-stages}
\end{figure*}

\paragraph{Pretraining time and memory.}
Post-replacement update speed and memory savings increase with the number of replaced heads (Figure~\ref{fig:systems-across-stages}a).
Updates are faster at every batch size from 2 to 8.
Table~\ref{tab:fixed-token-controls-main} reports full training-path times, including the initial training with ordinary attention and intervention.

\paragraph{Longer pretraining contexts.}
Without retuning, the perplexity penalty decreases from 4K to 16K at both rates (Table~\ref{tab:main-results}a).
Updates remain faster and use less peak memory at 8K and 16K, although their measured speedups are smaller than at 4K. \ourapproach{} is applied with models trained and patterns fitted separately at each context length.

\paragraph{Downstream adaptation.}
Across SST-2, BoolQ, and QuALITY, every absolute mean accuracy change is below 0.8 percentage points, with variation measured over three finetuning seeds (Table~\ref{tab:main-results}b).
QuALITY provides a long-input finetuning workload, although accuracy with ordinary attention is close to the four-choice baseline.
Finetuning speed depends on input length: SST-2 slows down, BoolQ approaches parity, and long-input QuALITY reaches 1.118$\times$ faster updates at 50\% replacement (Figure~\ref{fig:systems-across-stages}b).
Replacing heads during finetuning does not improve the quality--speed trade-off; Appendix~\ref{app:downstream-details} reports the results and controller costs.

\noindent\begin{minipage}{\textwidth}
\begin{wrapfigure}{r}{0.5\textwidth}
    \centering
    \includegraphics[width=\linewidth]{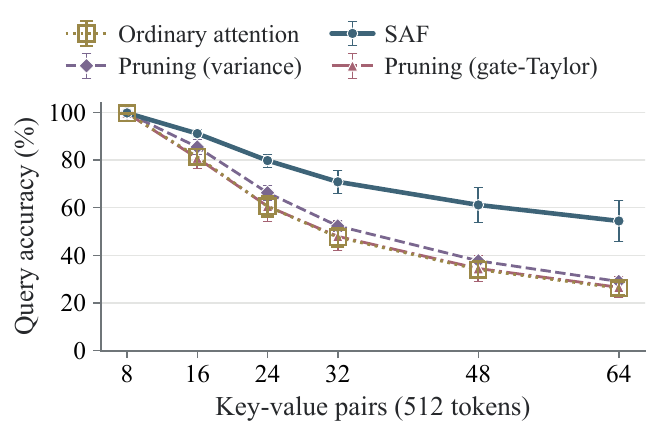}
    \caption{MQAR accuracy for later-intervention, matched-time 124M checkpoints.
    Models train on eight pairs at 512 tokens.
    (The ordinary-attention and gate-Taylor pruning curves nearly overlap.)}
    \label{fig:mqar-generalisation}
    \vspace{-0.2cm}
\end{wrapfigure}

\paragraph{Associative recall.}
Retaining fixed-pattern token mixing improves generalisation to more associations within the same context.
MQAR requires retrieving values paired with earlier keys \citep{ICLR2024_448fc91f}.
All four 124M models adapt on 512-token sequences with 8 pairs, reaching over 99.5\% mean accuracy.
At 64 pairs, \ourapproach{} retains 54.4\% accuracy versus 26.2--29.0\% for ordinary attention and pruning (Figure~\ref{fig:mqar-generalisation}).
It outperforms pruning controls in every seed at 24, 32, 48, and 64 pairs.
Length-only tests favour pruning, and immediate replacement in a task-trained ordinary-attention model does not give the same consistent benefit (Appendix~\ref{app:associative-recall}).
In the midpoint, matched-token controls, frozen mean also gives higher mean accuracy than ordinary attention at 16--64 pairs (Table~\ref{tab:pattern-controls}b).

At a fixed input length, increasing the number of key-value pairs tests generalisation to more associations rather than to longer sequences.
The higher MQAR accuracy indicates that these heads remain useful for retrieving key-value associations even when their attention weights are fixed. 
\end{minipage}

\paragraph{Causal prefill.}
For the 124M 16K RoPE checkpoints, 25\% and 50\% replacement accelerate causal prefill by 1.09--1.12$\times$ and 1.20--1.24$\times$ across input lengths at $B=64$.
Prefill is also faster across the 4K batch-size sweep (Figure~\ref{fig:systems-across-stages}c,d).
Appendix~\ref{app:kernel-details} gives absolute latencies and batch-one measurements, where launch overhead can outweigh the savings.
The computational benefit extends beyond pretraining: the same fixed patterns support faster long-input finetuning and causal prefill without being fitted again for these stages.

\noindent\begin{minipage}{\textwidth}
\begin{wrapfigure}{R}{0.6\textwidth}
\vspace{-0.6cm}
    \centering
    \includegraphics[width=\linewidth]{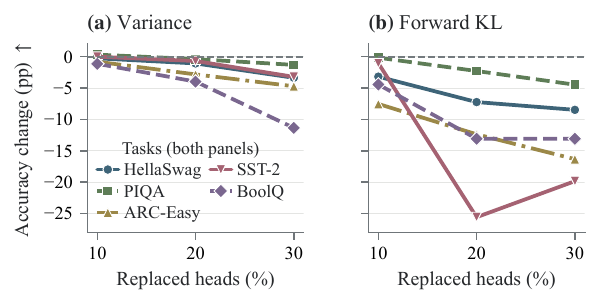}
    \vspace{-0.8cm}
    \caption{Qwen3-4B zero-shot accuracy changes (pp) relative to the ordinary-attention model.}
    \label{fig:qwen-pretrained-tasks}
\end{wrapfigure}

\paragraph{Zero-shot task accuracy in Qwen3-4B.}
Without further training, variance selection gives higher accuracy than forward KL on all tasks at 10\%, 20\%, and 30\% replacement (Figure~\ref{fig:qwen-pretrained-tasks}).
At 10\%, accuracy changes range from $-1.13$ to $+0.33$ percentage points relative to ordinary attention, averaging $-0.37$ points.
Higher replacement rates give larger mean accuracy losses.
Appendix~\ref{app:pretrained-static-details} reports per-task accuracies, selector analysis, and a separate GQA prefill benchmark.

\end{minipage}

\subsection{Does \ourapproach{} transfer to a larger model?}
\label{sec:scale-up-results}

At 1B and matched 19.667B training tokens, 25\% replacement increases perplexity by 0.507\% and changes mean finetuning accuracy by less than 0.72 percentage points per task (Table~\ref{tab:scale-up-results}).
At 50\%, perplexity increases by 2.027\% with larger update savings.
At 25\%, \ourapproach{} has slightly lower perplexity than Gate-Taylor pruning; mean task accuracies differ by less than 0.20 points, with no consistent advantage across finetuning seeds.
These comparisons use one pretraining seed and three finetuning seeds; QuALITY accuracy remains close to the four-choice baseline.

\begin{table}[h]
    \centering
    \small
    \setlength{\tabcolsep}{5pt}
    \caption{1B quality at 8K context and 19.667B training tokens. PPL uses one pretraining seed; task accuracies (\%) show mean $\pm$ SE over three finetuning seeds.}
    \label{tab:scale-up-results}
    \begin{tabular}{@{}lrrrr@{}}
        \toprule
        & & \multicolumn{3}{c}{Finetuning accuracy (\%) $\uparrow$} \\
        \cmidrule(l){3-5}
        Model & PPL $\downarrow$ & SST-2 & BoolQ & QuALITY \\
        \midrule
        Ordinary attention & 12.446 & $91.44\pm0.25$ & $75.98\pm0.24$ & $27.44\pm1.25$ \\
        \ourapproach{} 25\% & 12.509 & $91.63\pm0.35$ & $75.27\pm1.45$ & $27.60\pm0.21$ \\
        Pruning 25\% (gate-Taylor) & 12.529 & $91.70\pm0.34$ & $75.46\pm1.19$ & $27.66\pm0.71$ \\
        \ourapproach{} 50\% & 12.698 & $90.98\pm0.43$ & $75.68\pm0.31$ & $25.12\pm0.75$ \\
        \bottomrule
    \end{tabular}
\end{table}

With distributed communication on four GH200 GPUs, the 25\% and 50\% checkpoints give update speedups of 1.068$\times$ and 1.167$\times$, respectively.
These post-replacement measurements include gradient accumulation, optimisation, and all-reduce; Appendix~\ref{app:scale-up-details} reports separate single-GPU memory and finetuning-time measurements.
The 25\% setting reduces update time by 6.4\%, so applying this measured saving to a post-replacement budget of one million GPU-hours would save approximately 64,000 GPU-hours.
With midpoint replacement, the corresponding estimate is 3.2\% of the full update budget before the one-off intervention cost; the absolute saving grows with the training budget.

\section{Conclusion}
\label{sec:conclusion}

We have studied when attention heads can use fixed causal patterns during language-model training.
Controlled comparisons identify \fullourapproach{} (\ourapproach{}): variance selection and fitted mean patterns with compact storage and fused execution.
At 124M, freezing fitted patterns gives nearly the same perplexity as continued pattern learning, with faster updates.
At 25\% replacement, perplexity increases remain below 1\% at matched tokens, with post-replacement updates 1.056$\times$ faster at 124M and 4K, and 1.068$\times$ faster at 1B and 8K on four GPUs.
The models also support faster long-input finetuning and causal prefill.
At 124M, retaining token mixing improves matched-token perplexity over same-head pruning.
After associative-recall adaptation, it also improves generalisation to more key-value pairs at a fixed length over both pruning controls.
These findings give a practical recipe for reducing input-dependent attention computation while retaining useful token mixing.

\clearpage
\ifdefined\paperpreprintversion
\else
\section*{AI Use Statement}

In this work, we used generative AI tools to assist with method implementation and to refine experiments initially designed by the authors.
We have not used generative AI tools to formulate mathematical claims, develop or write proofs, propose or refine hypotheses, interpret results, or determine the manuscript structure.
Additionally, we used generative AI tools for code development and debugging, polishing text, and preparing result plots.
We take responsibility for the final content of this work, including text, code, analyses and claims.

\fi
\section*{Reproducibility Statement}

Section~\ref{sec:method} and Appendices~\ref{app:method-details}--\ref{app:recipe-analysis} detail the methods, assumptions, and derivations.
Section~\ref{sec:experimental-setup} and Appendices~\ref{app:details}--\ref{app:kernel-details} provide the experimental configurations, seeds, evaluation and measurement protocols, and run records.
Appendix~\ref{app:resources} lists datasets, pretrained models, and licences.
\ifdefined\paperpreprintversion
The codes, data split and model checkpoints are available at \url{https://github.com/waylonli/Selective-Attention-Freezing}.
\else
The codes, data split, and model checkpoints will be made publicly available upon acceptance.
\fi

\ifdefined\paperpreprintversion
\section*{Acknowledgements}

The authors acknowledge the use of resources provided by the Isambard-AI National AI Research Resource (AIRR).
Isambard-AI is operated by the University of Bristol and is funded by the UK Government's Department for Science, Innovation and Technology (DSIT) via UK Research and Innovation; and the Science and Technology Facilities Council [ST/AIRR/I-A-I/1023]~\citep{mcintoshsmith2024isambardaileadershipclasssupercomputer}.

\fi

\bibliography{iclr2027_conference}
\bibliographystyle{iclr2027_conference}

\clearpage
\appendix
\section{Additional Method Details}
\label{app:method-details}

We give the attention notation and calibration protocol, followed by the complete \ourapproach{} algorithm and definitions of the alternative selectors and schedules.

\subsection{Attention notation and calibration}
\label{app:attention-notation}

Consider a model with $L$ attention layers and $H$ query heads per layer, giving $LH$ heads in total.
Let $\ell\in\{1,\ldots,L\}$ index layers, $h\in\{1,\ldots,H\}$ index heads, $T$ denote sequence length, $d$ denote the hidden dimension, and $d_h$ denote the head dimension.
For an input sequence, $\mathbf{R}_{\ell}\in\mathbb{R}^{T\times d}$ denotes the residual stream entering layer $\ell$.
The attention input is $\mathbf{X}_{\ell}=\operatorname{LN}_{\ell}(\mathbf{R}_{\ell})$, where $\operatorname{LN}_{\ell}$ is the pre-attention layer normalisation \citep{pmlr-v119-xiong20b}.
The projections are $\mathbf{Q}_{\ell h}=\mathbf{X}_{\ell}\mathbf{W}^{Q}_{\ell h}$, $\mathbf{K}_{\ell h}=\mathbf{X}_{\ell}\mathbf{W}^{K}_{\ell h}$, and $\mathbf{V}_{\ell h}=\mathbf{X}_{\ell}\mathbf{W}^{V}_{\ell h}$, with projection matrices in $\mathbb{R}^{d\times d_h}$ and outputs in $\mathbb{R}^{T\times d_h}$.
Rotary position embeddings, when used, act on $\mathbf{Q}$ and $\mathbf{K}$ before the dot product \citep{su2023roformer}.
Standard causal attention maps $\mathbf{X}_{\ell}$ to
\begin{align}
    \mathbf{S}_{\ell h}(\mathbf{X}_{\ell})
    &= \frac{\mathbf{Q}_{\ell h}\mathbf{K}_{\ell h}^{\top}}{\sqrt{d_h}}, \\
    \attn_{\ell h}(\mathbf{X}_{\ell})
    &= \operatorname{softmax}_{\mathrm{causal}}\!\left(\mathbf{S}_{\ell h}(\mathbf{X}_{\ell})\right), \\
    \mathbf{Z}_{\ell h}(\mathbf{X}_{\ell})
    &= \attn_{\ell h}(\mathbf{X}_{\ell})\mathbf{V}_{\ell h}.
    \label{eq:standard-head}
\end{align}
Writing $\mathbf{W}^{O}_{\ell h}\in\mathbb{R}^{d_h\times d}$ for the output-projection block of head $h$, its contribution to the residual stream is $\mathbf{C}_{\ell h}=\mathbf{Z}_{\ell h}\mathbf{W}^{O}_{\ell h}$.
Let $i,j\in\{1,\ldots,T\}$ index query and key positions, respectively.
The operator $\operatorname{softmax}_{\mathrm{causal}}$ masks future keys and normalises each query row, so $\attn_{\ell h}(i,j)=0$ for $j>i$, $\attn_{\ell h}(i,j)\geq0$, and $\sum_{j=1}^{i}\attn_{\ell h}(i,j)=1$.
Therefore, the attention matrix is lower triangular (including the diagonal) and row-stochastic.

We measure attention at the replacement checkpoint in evaluation mode, keeping its weights fixed.
The original 124M and 1B pretraining runs sample calibration sequences from the training corpus; the additional three-seed controls use a held-out calibration partition.
Both protocols use calibration data separate from final reporting and leave the training-data sampler unchanged; all head scores and pattern targets are computed before replacement.
Appendices~\ref{app:controlled-pretraining-details}, \ref{app:fixed-token-controls}, and~\ref{app:scale-up-details} give the data partitions and calibration budgets.

\subsection{Training procedure}
\label{app:training-algorithm}

Algorithm~\ref{alg:calibration} specifies the one-time \ourapproach{} recipe.
The model first completes $t_\star$ of $M$ optimiser updates, then measures attention on $\mathcal{D}_{\mathrm{cal}}$ using its current weights.
Each $\mathcal{D}_t$ contains the micro-batches for one complete training update; \textsc{LMUpdate} applies the language-model objective, prescribed learning rate, and optimiser state $\omega$.
Calibration leaves model weights and the training-data order unchanged.
The calibration partitions differ between the original pretraining runs and the supplementary controls, as specified in Appendices~\ref{app:controlled-pretraining-details}, \ref{app:fixed-token-controls}, and~\ref{app:scale-up-details}.

\begin{algorithm}[t]
    \caption{\fullourapproach{} (\ourapproach{}) with one-time replacement.}
    \label{alg:calibration}
    \begin{algorithmic}[1]
        \Require Model $f_\theta$, optimiser state $\omega$, training batches $\{\mathcal{D}_t\}_{t=1}^{M}$,
        \Statex \hspace{\algorithmicindent}calibration sequences $\mathcal{D}_{\mathrm{cal}}$ of length $T$
        \Require Replacement update step $t_\star$, rate $r$, fit steps $J$, fit learning rate $\gamma$, tolerance $\varepsilon_{\mathrm{fit}}$
        \Ensure Trained $f_\theta$, replaced heads $\mathcal{F}$, fixed pattern buffers $\mathcal{B}$
        \For{$t=1,\ldots,t_\star$}
            \State $(\theta,\omega)\gets\Call{LMUpdate}{\theta,\omega,\mathcal{D}_t,t}$
        \EndFor
        \State Measure $\widehat{\attn}_{\ell h}$ (Section~\ref{sec:calibration-selection}) and $s^{\mathrm{var}}_{\ell h}$ (Equation~\ref{eq:variance-score}) on $\mathcal{D}_{\mathrm{cal}}$ in evaluation mode, holding $\theta$ fixed
        \State $\pi\gets$ all head indices sorted by increasing $s^{\mathrm{var}}_{\ell h}$
        \State $k\gets\operatorname{round}(rLH)$; $\mathcal{F}\gets\varnothing$; $\mathcal{B}\gets\varnothing$
        \For{head $u=(\ell,h)$ in $\pi$}
            \If{$|\mathcal{F}|=k$}
                \State \textbf{break}
            \EndIf
            \State Floor causal entries of $\widehat{\attn}_u$ at $10^{-9}$, then row-normalise to obtain $\prior_u$
            \State Compute $\eta_{\mathrm{abs}},\eta_{\mathrm{rel}}$ from $\prior_u$ (Equation~\ref{eq:fit-objective})
            \State $(\alpha_u,\rho_u)\gets\Call{Fit}{\eta_{\mathrm{abs}},\eta_{\mathrm{rel}},J,\gamma}$
            \State Compute $Z_u$ and $\widehat{\prior}_u$ as in Equation~\ref{eq:compact-prior}
            \State $\kappa_u\gets T^{-1}\sum_{i=1}^{T}\mathrm{KL}\!\left(\prior_u[i,1{:}i]\,\|\,\widehat{\prior}_u[i,1{:}i]\right)$
            \If{$\kappa_u\leq\varepsilon_{\mathrm{fit}}$}
                \State $\mathcal{F}\gets\mathcal{F}\cup\{u\}$; $\mathcal{B}[u]\gets(\alpha_u,\rho_u,\log Z_u)$
            \EndIf
        \EndFor
        \State Verify that $|\mathcal{F}|=k$ for the requested replacement rate
        \State Install $\mathcal{B}$ as fixed attention patterns for $\mathcal{F}$; retain value/output weights and optimiser state
        \State Resume training mode with $\mathcal{F}$ and $\mathcal{B}$ fixed
        \For{$t=t_\star+1,\ldots,M$}
            \State $(\theta,\omega)\gets\Call{LMUpdate}{\theta,\omega,\mathcal{D}_t,t}$ using mixed-head attention
        \EndFor
        \State \Return $f_\theta,\mathcal{F},\mathcal{B}$
    \end{algorithmic}
\end{algorithm}

\paragraph{Selection, fitting, and installation.}
The algorithm considers candidates in increasing variance order, skips failed compact fits, and continues until $k$ heads pass the fitting threshold or the candidates are exhausted.
The requested rate is attained only if $k$ fits pass. Variance determines candidate order, while $\varepsilon_{\mathrm{fit}}$ bounds the compact-fitting error.
It represents batched fitting as a headwise loop; the model receives no training updates between candidate fits.
Comparisons that prescribe identical head sets instead fit only the specified heads and reject a configuration if any fit fails, without substituting other heads.
The additional fixed-token controls use this rule for variance-selected means (Appendix~\ref{app:fixed-token-controls}).
\textsc{Fit} initialises $\alpha(j)=0$ and $\rho(\delta)=\log\max\{\eta_{\mathrm{rel}}(\delta)/(T-\delta),10^{-9}\}$, then applies $J=400$ Adam steps at learning rate $\gamma=0.05$ to Equation~\ref{eq:fit-objective}, without early stopping.
The acceptance threshold is $\varepsilon_{\mathrm{fit}}=0.2$ nats.
Computing $\widehat{\prior}$ and its KL in the algorithm denotes evaluation of the fitted distribution; the FFT fitter obtains normalisers and KL from sufficient statistics without materialising that matrix.
After installation, the fixed pattern parameters receive no language-model gradients; ordinary attention and the remaining model parameters, including value and output projections, continue to train.

\paragraph{Schedule variants.}
The default uses one event near the training midpoint; earlier or later one-time replacement changes $t_\star$.
For a schedule $\{(t_k,\mathcal{H}_k)\}_{k=1}^{K}$, the same calibration, fitting, and installation operations apply at each event to newly proposed heads $\mathcal{H}_k$.
The gradual comparison fixes the ordered final head set and installs it in five groups, fitting each group's patterns at its replacement event.
Adaptive controllers determine events from validation or training loss and restrict proposals to heads that still use ordinary attention.
The validation-budget controller accepts an increment only if its paired loss change fits within the cumulative budget; the train-loss-triggered controller waits for a plateau and subsequent recovery, and can undo an unsuccessful increment.
Previously installed patterns remain fixed while new candidates are measured.
These alternatives modify the event schedule and acceptance policy, rather than making recalibration part of every training update; Appendix~\ref{app:selector-schedule-definitions} gives their detailed settings.

\subsection{Alternative selector and schedule definitions}
\label{app:selector-schedule-definitions}

Section~\ref{sec:calibration-selection} defines the attention-variance and forward-KL scores.
Appendix~\ref{app:head-selection-analysis} reports results for the gradient and output-aware scores defined below, including preliminary gradient comparisons, one-head diagnostics, and continuation experiments.
For the gradient alternative, let $t$ index optimiser updates, let $\mathcal{L}_t$ be the current minibatch language-model loss, and define
\begin{equation}
    q_{\ell h}^{(t)}
    =\lVert\nabla_{\mathbf{W}^{Q}_{\ell h}}\mathcal{L}_t\rVert_F
    +\lVert\nabla_{\mathbf{W}^{K}_{\ell h}}\mathcal{L}_t\rVert_F,
    \qquad
    \overline q^{(t)}=(LH)^{-1}\sum_{\ell',h'}q_{\ell'h'}^{(t)}.
\end{equation}
We remove the scale shared across heads at update $t$ using
\begin{equation}
    s_{\ell h}^{\mathrm{grad}}(t)=\beta s_{\ell h}^{\mathrm{grad}}(t-1)
    +(1-\beta)\frac{q_{\ell h}^{(t)}}{\overline q^{(t)}},
    \qquad \beta=0.9,
    \label{eq:gradient-score}
\end{equation}
initialised by $s_{\ell h}^{\mathrm{grad}}(1)=q_{\ell h}^{(1)}/\overline q^{(1)}$.
The symmetric rank combination is $s_{\ell h}^{\mathrm{sum}}=\operatorname{rank}(s_{\ell h}^{\mathrm{var}})+\operatorname{rank}(s_{\ell h}^{\mathrm{grad}})$, with rank zero assigned to the smallest value across all $LH$ heads.
The asymmetric gradient-veto rule orders heads by $s_{\ell h}^{\mathrm{var}}$ but places heads with above-median $s_{\ell h}^{\mathrm{grad}}$ after the remaining candidates.

The output-aware scores use a separate set of $N_f$ held-out sequences indexed by $m$ and query positions $\mathcal{I}$ sampled evenly from the latter half of each sequence.
The symbols $\mathbf{R}_{\ell,m}$ and $\mathbf{X}_{\ell,m}$ denote the residual input and normalised attention input from Appendix~\ref{app:attention-notation} on sequence $m$.
Define $\mathbf{V}_{\ell h}^{(m)}=\mathbf{X}_{\ell,m}\mathbf{W}_{\ell h}^{V}$, $\attn_{\ell h}^{(m)}=\attn_{\ell h}(\mathbf{X}_{\ell,m})$, and $\mathbf{C}_{\ell h}^{(m)}=\attn_{\ell h}^{(m)}\mathbf{V}_{\ell h}^{(m)}\mathbf{W}_{\ell h}^{O}$.
The vector $\mathbf{c}_{\ell h}$ concatenates $\mathbf{C}_{\ell h}^{(m)}[i]$ over $m$ and $i\in\mathcal{I}$.
The projected-output magnitude is $s_{\ell h}^{\mathrm{mag}}=\lVert\mathbf{c}_{\ell h}\rVert_2/\sqrt{\dim(\mathbf{c}_{\ell h})}$, and within-layer redundancy is
\begin{equation}
    s_{\ell h}^{\mathrm{red}}=-\max_{h'\ne h}
    \frac{\langle\mathbf{c}_{\ell h},\mathbf{c}_{\ell h'}\rangle}
    {\lVert\mathbf{c}_{\ell h}\rVert_2\lVert\mathbf{c}_{\ell h'}\rVert_2}.
    \label{eq:redundancy-score}
\end{equation}
The maximum ranges over heads in the same layer, and the minus sign gives higher replacement priority to highly redundant heads.

Let $\widetilde{\mathbf{C}}_{\ell h}^{(m)}=\widehat{\attn}_{\ell h}\mathbf{V}_{\ell h}^{(m)}\mathbf{W}_{\ell h}^{O}$, $\mathbf{O}_{\ell}^{(m)}=\sum_h\mathbf{C}_{\ell h}^{(m)}$, and $\mathbf{Y}_{\ell}^{(m)}=\mathbf{R}_{\ell,m}+\mathbf{O}_{\ell}^{(m)}$.
Replacing only head $h$ gives $\widetilde{\mathbf{Y}}_{\ell h}^{(m)}=\mathbf{Y}_{\ell}^{(m)}-\mathbf{C}_{\ell h}^{(m)}+\widetilde{\mathbf{C}}_{\ell h}^{(m)}$.
The two replacement-influence scores are
\begin{align}
    s_{\ell h}^{\mathrm{cos}}
    &=\mathbb{E}_{m,i}\!\left[1-\cos\!\left(\mathbf{Y}_{\ell}^{(m)}[i],
      \widetilde{\mathbf{Y}}_{\ell h}^{(m)}[i]\right)\right], \\
    s_{\ell h}^{\mathrm{rel}}
    &=\mathbb{E}_{m,i}\!\left[
      \frac{\lVert\widetilde{\mathbf{C}}_{\ell h}^{(m)}[i]-\mathbf{C}_{\ell h}^{(m)}[i]\rVert_2}
      {\max\!\left(\lVert\mathbf{O}_{\ell}^{(m)}[i]\rVert_2,10^{-8}\right)}\right].
    \label{eq:output-influence-scores}
\end{align}
The local diagnostic compares these scores with the paired held-out NLL change caused by replacing one head with its fitted mean pattern.

For a schedule $\{(t_k,\mathcal{H}_k)\}_{k=1}^{K}$, heads $\mathcal{H}_k$ are replaced at update $t_k$.
Let $\mathcal{F}=\bigcup_{k=1}^{K}\mathcal{H}_k$ be the final selected set and $\tau=t/M$ the intervention fraction of an $M$-update run.
The one-time schedule replaces all heads in $\mathcal{F}$ at update $t$, while the gradual schedule partitions $\mathcal{F}$ into five nearly equal groups applied at evenly spaced updates.
The validation-budget schedule accepts small groups in score order while their cumulative paired validation-perplexity cost remains below a specified relative budget.
The train-loss-triggered schedule proposes a group when smoothed training loss satisfies a plateau test and waits for recovery before the next proposal.
Its parameter $z$ is the standard-deviation multiplier in the plateau, statistical-power, and recovery tests; larger $z$ permits an earlier trigger.
The adaptive schedules determine their final rates through their stopping rules and configured caps.

\FloatBarrier
\section{Understanding the Selected Recipe}
\label{app:recipe-analysis}

Our controlled study selects the post-softmax mean as the fixed pattern and attention variance as the head-selection score.
Corpus-averaged attention has also been used as an input-independent replacement in prior probing work \citep{hassid2022much}.
Both choices follow from the same objective for approximation by a fixed matrix: the mean gives the best fixed pattern, while the variance measures how accurately each head can be approximated by one.

\subsection{Fixed-pattern reconstruction}

Fix one layer and head, suppress the indices $\ell,h$, and write $\attn(\mathbf{X})$ for its attention matrix on input $\mathbf{X}$ and $\prior$ for its input-independent replacement.
Let $\lVert\cdot\rVert_F$ denote the Frobenius norm.
For a fixed replacement pattern $\prior$, define the expected reconstruction error per matrix entry as
\begin{equation}
    \mathcal{R}(\prior)
    = \frac{1}{T^2}\mathbb{E}_{\mathbf{X}}
      \left[\lVert\attn(\mathbf{X})-\prior\rVert_F^2\right].
    \label{eq:routing-risk}
\end{equation}
Writing $\meanattn=\mathbb{E}_{\mathbf{X}}[\attn(\mathbf{X})]$ for the population mean gives
\begin{equation}
    \mathcal{R}(\prior)
    = \mathcal{R}(\meanattn)
      + \frac{1}{T^2}\lVert\prior-\meanattn\rVert_F^2.
    \label{eq:routing-bias-variance}
\end{equation}
Because $\meanattn$ is an average of causal row-stochastic matrices, it is itself a valid attention matrix and minimises this reconstruction error.
The minimum value $\mathcal{R}(\meanattn)$ is the population attention variance per matrix entry, which $s_{\ell h}^{\mathrm{var}}$ in Equation~\ref{eq:variance-score} estimates from calibration inputs.
The same objective therefore determines both what fixed pattern to use and which heads are easiest to replace.

This argument applies to attention reconstruction rather than directly to language-model loss.
The empirical mean $\widehat{\attn}_{\ell h}$ only estimates the population mean $\meanattn$, while language-model loss can also depend on value vectors, output projections, interactions between heads, and further training after replacement.
This reconstruction analysis nevertheless shows that input-independent sampling has no inherent advantage under this objective: a sampled pattern introduces variation without conditioning that variation on the current input.

Probability geometry provides a complementary view of the two mean patterns.
These centroid properties are instances of the broader connection between Bregman divergences and their corresponding mean representations \citep{banerjee2005clustering}.
For one valid causal row, let $\mathbf{a}(\mathbf{X})$ be the random attention vector and $\mathbf{p}$ a fixed probability vector on the same support.
The two directions of KL divergence select different fixed summaries:
\begin{align}
    \argmin_{\mathbf{p}}\;\mathbb{E}_{\mathbf{X}}
    \left[\KL\!\left(\mathbf{a}(\mathbf{X})\,\|\,\mathbf{p}\right)\right]
    &= \mathbb{E}_{\mathbf{X}}[\mathbf{a}(\mathbf{X})],
    \label{eq:forward-kl-barycenter} \\
    \argmin_{\mathbf{p}}\;\mathbb{E}_{\mathbf{X}}
    \left[\KL\!\left(\mathbf{p}\,\|\,\mathbf{a}(\mathbf{X})\right)\right]
    &= \frac{\exp(\mathbb{E}_{\mathbf{X}}[\log\mathbf{a}(\mathbf{X})])}
            {\sum_j\exp(\mathbb{E}_{\mathbf{X}}[\log a_j(\mathbf{X})])}.
    \label{eq:reverse-kl-barycenter}
\end{align}
The first identity follows from the cross-entropy term, and the second from a Lagrange multiplier enforcing $\sum_jp_j=1$.
If $\mathbf{a}=\operatorname{softmax}(\mathbf{s})$, the shared row-normalisation term cancels in the second expression, which becomes $\operatorname{softmax}(\mathbb{E}[\mathbf{s}])$.
Thus, the post-softmax and sharp means are respectively the forward- and reverse-KL barycentres.

We analyse fixed-pattern distances to check whether these geometries are consistent with the observed ordering, without treating them as an explanation of language-model loss.
For each fixed pattern, we compare its stored matrix $\prior$ with the measured post-softmax mean on the same nested head subset. In these empirical comparisons, $\meanattn$ denotes this measured mean before compact fitting.
Table~\ref{tab:prior-geometry-correlation} ranks the five patterns in Figure~\ref{fig:prior-quality} separately within each replacement rate, so that the shared effect of replacement rate does not inflate the association with final perplexity.
Squared distance has a stratified Spearman correlation of 0.950 with perplexity increase, compared with 0.625 for $\KL(\meanattn\|\prior)$ and 0.900 for both squared Hellinger distance and total variation.
These correlations are descriptive because each within-rate comparison contains only five patterns and the post-softmax mean has zero squared distance by construction.

\begin{table}[t]
    \centering
    \small
    \caption{Fixed-pattern geometry: within-rate correlation between distance from the post-softmax mean and final perplexity increase.}
    \label{tab:prior-geometry-correlation}
    \begin{tabular}{lrrrrr}
        \toprule
        Distance from mean & 10\% & 25\% & 50\% & 75\% & Stratified \\
        \midrule
        Squared error       & 0.900 & 1.000 & 1.000 & 0.900 & 0.950 \\
        $\KL(\meanattn\|\prior)$ & 0.600 & 0.600 & 0.600 & 0.700 & 0.625 \\
        Squared Hellinger   & 0.800 & 0.900 & 0.900 & 1.000 & 0.900 \\
        Total variation     & 0.800 & 0.900 & 0.900 & 1.000 & 0.900 \\
        \bottomrule
    \end{tabular}
\end{table}

\FloatBarrier
\subsection{Selecting heads for continued pretraining}
\label{app:head-selection-analysis}

Head selection can be evaluated by the immediate cost of replacing one head or by the final loss after replacing a complete set and continuing training.
These two evaluations need not agree because the latter also reflects interactions among selected heads and subsequent adaptation.

Preliminary 124M comparisons at 1K context use dense post-softmax means, 25\% midpoint replacement, and one run per selector.
Q/K gradient selection gives similar final perplexity to variance (26.558 versus 26.568), while the rank sum and gradient-veto rule increase it to 26.627 and 26.607, respectively; training with ordinary attention gives 26.382.
These dense-pattern results are separate from the compact-pattern comparisons below.

At the 4K midpoint, we test whether output-aware scores predict immediate replacement cost more accurately than attention variance.
We fit a compact pattern to the measured post-softmax mean of each of 144 heads and measure the paired NLL change on held-out windows when replacing one head at a time.
The score comparisons use the 118 heads that pass the compact-fitting acceptance threshold.
Table~\ref{tab:selector-diagnostic} compares attention variance with four output-aware scores and reports their association with this local cost and the mean cost over their lowest-risk 25\% and 50\% subsets.

\begin{table*}[t]
    \centering
    \small
    \caption{Immediate one-head diagnostics at the 4K RoPE midpoint.
    Spearman correlates scores with NLL changes; mean costs are relative perplexity increases.}
    \label{tab:selector-diagnostic}
    \begin{tabular}{lrrr}
        \toprule
        Risk score & Spearman $\uparrow$ & 25\% mean cost $\downarrow$ & 50\% mean cost $\downarrow$ \\
        \midrule
        Attention variance, $s^{\mathrm{var}}$ & 0.723 & 0.058\% & \textbf{0.115\%} \\
        Residual cosine, $s^{\mathrm{cos}}$ & \textbf{0.810} & \textbf{0.033\%} & 0.121\% \\
        Relative attention-output change, $s^{\mathrm{rel}}$ & 0.703 & 0.052\% & 0.150\% \\
        Projected-output magnitude, $s^{\mathrm{mag}}$ & 0.654 & 0.166\% & 0.129\% \\
        Projected-output redundancy, $s^{\mathrm{red}}$ & $-0.013$ & 0.320\% & 0.367\% \\
        \bottomrule
    \end{tabular}
\end{table*}

Residual cosine predicts the immediate one-head cost more closely and selects a 25\% set with a smaller joint perturbation than variance.
The sets share 23 of 36 heads, and their immediate PPL increases on 16 new common validation windows are 4.136\% and 5.092\%, respectively.
After replacing both sets from the same update-2,500 checkpoint and continuing for the same 2,500 updates, however, their final PPL increases are 0.792\% and 0.665\%.
Resampling the 598 matched 4K evaluation windows gives a residual-cosine-minus-variance PPL difference of 0.126\% (95\% interval: 0.102--0.150\%).
A small local perturbation from replacing one head therefore does not necessarily predict the final result after jointly replacing many heads and continuing training.
The discrepancy may reflect interactions among selected heads, subsequent adaptation, or both.

We next compare the two input-dependence scores, attention variance and row-wise forward KL, after adaptation is complete.
For every context length and replacement rate, both methods select a complete head set at the midpoint and undergo the same 2,500 continuation updates.

\begin{table}[t]
    \centering
    \small
    \caption{Final perplexity increase after joint head selection, midpoint replacement, and 2,500 continuation updates.}
    \label{tab:kl-continuation-selector}
    \begin{tabular}{rrrrr}
        \toprule
        Context & Replaced & Variance & Forward KL & KL--variance \\
        & heads & $\Delta$PPL & $\Delta$PPL & (pp) \\
        \midrule
        4K  & 25\% & 0.665\% & 1.068\% & +0.402 \\
        4K  & 50\% & 2.649\% & 3.554\% & +0.905 \\
        8K  & 25\% & 0.557\% & 0.836\% & +0.279 \\
        8K  & 50\% & 2.191\% & 2.988\% & +0.797 \\
        16K & 25\% & 0.474\% & 0.876\% & +0.402 \\
        16K & 50\% & 1.802\% & 2.417\% & +0.615 \\
        \bottomrule
    \end{tabular}
\end{table}

Attention variance gives the lower final perplexity increase in all six matched settings.

\begin{figure}[t]
    \centering
    \includegraphics[width=\textwidth]{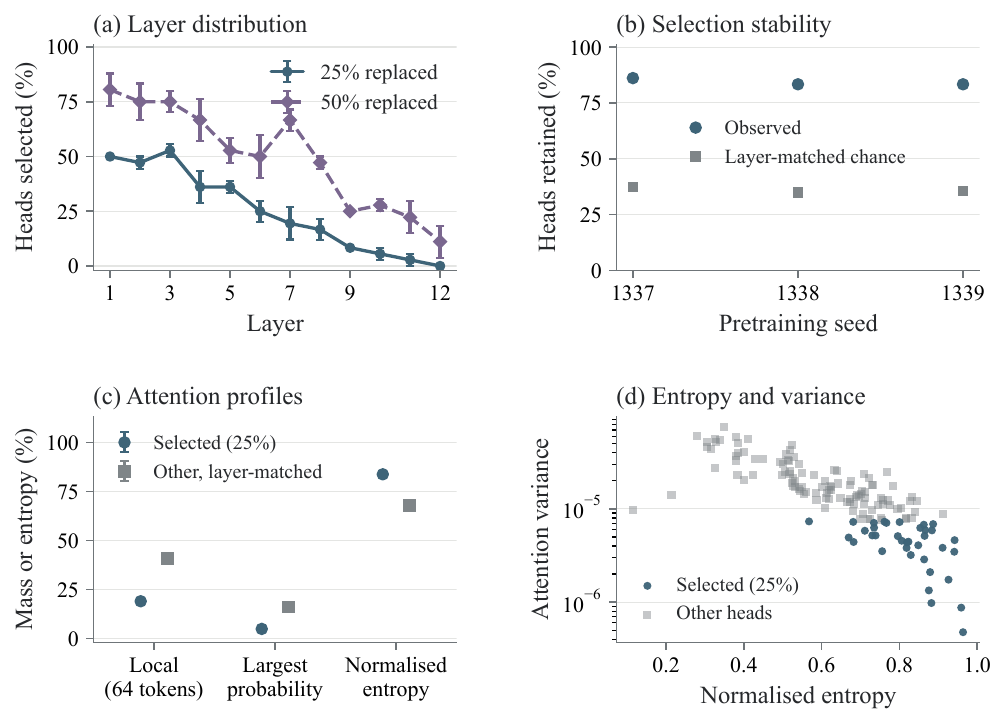}
    \caption{Selected-head diagnostics: (a) layer distribution; (b) quarter-to-midpoint retention; (c) midpoint attention profiles; (d) entropy and variance in the original one-head probe. Bars in (a,c) show mean $\pm$ SE over three seeds.}
    \label{fig:head-selection-profiles}
\end{figure}

The selected heads are concentrated in earlier layers, and this distribution is consistent across seeds.
In the three-seed 4K controls of Appendix~\ref{app:fixed-token-controls}, the first four layers contain 21--24 of the 36 heads selected at 25\% replacement (Figure~\ref{fig:head-selection-profiles}a).
Selection is also stable within each training run: the quarter-training and midpoint sets share 30--31 heads, compared with an expected 12.6--13.4 for independent selections with the same per-layer counts (Figure~\ref{fig:head-selection-profiles}b).
The recorded variance rankings have Spearman correlations of 0.959--0.962 between these stages.
We compare layer distributions across seeds and head identities within a seed, since head indices need not represent the same function in independently trained models.

To characterise the selected patterns, we evaluate the ordinary-attention checkpoints on 128 common, non-overlapping 4K validation windows and 64 evenly spaced query positions in the latter half of each window.
These inputs are separate from the final perplexity evaluation.
For a causal query row $\mathbf{p}=(p_1,\ldots,p_i)$, we measure local mass $\sum_{j=\max(1,i-63)}^{i}p_j$, the largest probability $\max_jp_j$, first-token mass $p_1$, and normalised entropy $-\sum_{j=1}^{i}p_j\log p_j/\log i$.
We average each quantity across inputs and query positions, then compare the 36 selected heads with unselected heads weighted to match their layer distribution.
The selected heads spread attention more broadly: their mean normalised entropy is 0.837 versus 0.678, and their mass on the most recent 64 tokens, including the current token, is 19.0\% versus 40.7\% (Figure~\ref{fig:head-selection-profiles}c).
Only 0.023\% of their attention falls on the first token on average, so first-token concentration does not characterise the selected set.
These measurements describe attention shape rather than assigning linguistic functions or establishing attention sinks.

Variance remains associated with local replacement cost after adjusting for layer and entropy.
We add entropy measurements to the original seed-1337 one-head diagnostic above, using its same checkpoint and 16 calibration windows, with the same 64 query positions as the profile analysis.
Among its 118 eligible heads, variance and normalised entropy have Spearman correlation $-0.798$.
The variance--replacement-cost correlation is 0.723 before adjustment, 0.733 after controlling for layer, and 0.543 after also controlling for entropy.
For these partial rank correlations, we regress the ranked variance and ranked one-head NLL change on layer indicators, additionally include ranked entropy in the latter comparison, and correlate the residuals.
Figure~\ref{fig:head-selection-profiles}d shows all 144 heads, highlighting the eligible 25\% selection.
Thus, the score favours diffuse attention but retains information about local replacement cost beyond layer and entropy; the continuation and pruning comparisons test whether retaining that attention as a fixed pattern benefits the final model.

A matched-input comparison with the final ordinary-attention checkpoints shows that rankings continue to develop after quarter-training.
Using the same 128 windows and sampled query rows at every stage, variance ranks correlate with their final ordering at 0.869--0.904 at quarter-training and 0.945--0.966 at the midpoint.
These later-query diagnostics complement the recorded full-matrix selection scores: substantial overlap is already present early, while the midpoint ordering is more representative of the final model.

\FloatBarrier
\subsection{Adaptation after replacement}

\begin{figure}[t]
    \centering
    \includegraphics[width=0.82\textwidth]{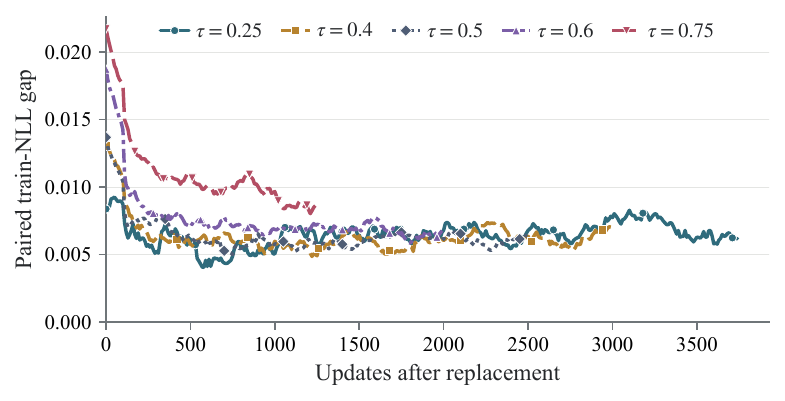}
    \caption{Paired train-loss gap after replacing 25\% of heads at training fraction $\tau$.}
    \label{fig:replacement-recovery}
\end{figure}

Replacement time introduces a second factor beyond the static reconstruction objective.
Earlier replacement leaves more updates for the model to adjust, while later replacement uses attention statistics from a more developed model.
In the replacement-time comparison, the immediate replacement cost increases from 2.72\% at $\tau=0.25$ to 7.99\% at $\tau=1$.
The observed training trajectories directly support this adaptation effect.
Figure~\ref{fig:replacement-recovery} subtracts the ordinary-attention run's minibatch loss from each replacement run at the same update and data order and applies a centred 210-update moving average.
Two hundred updates after replacement, the smoothed train-NLL gap is 0.0061 nats at $\tau=0.25$ and 0.0123 nats at $\tau=0.75$; by the final update, the corresponding gaps are 0.0062 and 0.0086 nats.
Continued training can therefore offset part of the replacement cost, while late replacement leaves less time for recovery.
The head diagnostics above show substantial selection overlap between quarter-training and the midpoint, alongside further changes towards the final ranking.
The timing comparison therefore concerns both the fixed approximation available at replacement and the subsequent adaptation, rather than simply whether the selected heads have stabilised.

\FloatBarrier
\section{Additional Experimental Details}
\label{app:details}

We first give the controlled pretraining configurations and pattern diagnostics, then compare replacement with pruning under matched budgets.
The remaining subsections cover 124M finetuning, 1B training and evaluation, and replacement in pretrained Qwen3-4B.

\FloatBarrier
\subsection{Controlled pretraining results}
\label{app:controlled-pretraining-details}

We train a 124M-parameter decoder with 12 Transformer layers, 12 heads per layer, hidden dimension $d=768$, and context length $T=4096$ on FineWeb-Edu \citep{penedo2024fineweb}.
We compare learned absolute position embeddings with RoPE \citep{su2023roformer}.
Each run performs 5,000 optimiser updates with 491,520 target tokens per update, for a total of 2.4576B target tokens.
This budget is approximately 20 tokens per parameter \citep{hoffmann2022chinchilla}.
The 4K quality runs use micro-batch size $B=2$ and 60 gradient-accumulation steps.
The 8K and 16K quality runs use micro-batch size one with 60 and 30 accumulation steps, respectively, preserving the tokens per update and total training budget.
The method-selection comparisons use seed 1337.
After selecting two operating points from the replacement-rate comparison, we repeat these configurations and their ordinary-attention baselines with seeds 1338 and 1339.

Replacement runs start from the same pretrained checkpoint, retaining the remaining training configuration and data order.
Pattern comparisons use the same selected heads, while rate comparisons use nested subsets of the same head ranking.
Unless replacement timing is being studied, heads are replaced halfway through training.
Every replacement run records the source checkpoint hash, parent checkpoint hash, training configuration, selected head list, fitted pattern, evaluation interval, and final checkpoint hash.
Replacement-time comparisons recompute the ordering at each parent checkpoint and use that ordering to pair mean and random patterns.

Table~\ref{tab:decision-points} summarises the controlled alternatives defined in Section~\ref{sec:method}.
We treat position encoding as a model control rather than a choice introduced by attention replacement.

\begin{table}[!htbp]
    \centering
    \caption{Method choices evaluated in controlled 124M pretraining.}
    \label{tab:decision-points}
    \small
    \setlength{\tabcolsep}{5pt}
    \begin{tabular}{@{}p{0.27\linewidth}p{0.66\linewidth}@{}}
        \toprule
        Method choice & Alternatives evaluated \\
        \midrule
        Fixed pattern & Post-softmax mean; sharp mean; Gaussian sample; Dirichlet sample; structured random \\
        \cmidrule(lr){1-2}
        Head selection & Attention variance; row-wise forward KL; Q/K gradient norm; rank combinations; projected-output magnitude and redundancy; attention-residual influence \\
        \cmidrule(lr){1-2}
        Amount and timing & Rates: 10\%, 25\%, 50\%, 75\%; times: 25\%, 40\%, 50\%, 60\%, 75\%, 100\%; one-time, gradual, validation-budget, and train-loss-triggered schedules \\
        \cmidrule(lr){1-2}
        Representation and execution & Exact dense or absolute-plus-relative pattern; reference or fused mixed-head execution \\
        \bottomrule
    \end{tabular}
\end{table}

Gaussian, Dirichlet, and sharp patterns use dense storage in the pattern-content comparison; the mean is tested in both dense and absolute-plus-relative forms, with the latter used for systems measurements.
Dirichlet concentration is scaled by 16, and structured-random patterns are reused across matched rates and replacement times.
Gradual replacement uses $K=5$ groups with the same final heads $\mathcal{F}$ and rate as one-time replacement; we also compare adaptive schedules at matched rates.
Controller settings are given in Appendix~\ref{app:selector-schedule-definitions}.

We additionally compare the selected recipe with head pruning and random head selection over three seeds at the same token budget.
Pruning uses the variance-selected head sets; random selection retains fitted means and matches the number of replaced heads in each layer.
These runs use $B=8$, 15 accumulation steps, and a shuffled non-overlapping data order, with their own ordinary-attention references (Appendix~\ref{app:fixed-token-controls}).

\paragraph{Optimisation and calibration.}
We use AdamW with a peak learning rate of $6\times10^{-4}$, linear warm-up over 500 updates, and cosine decay to $6\times10^{-5}$ at update 5,000.
The optimiser uses $\beta_1=0.9$, $\beta_2=0.95$, weight decay 0.1, and gradient clipping at norm 1; dropout is zero.
Calibration uses 16 micro-batches at the training context length: 32 sequences at 4K and 16 at 8K and 16K, sampled from the configured training-data interval with a separate random-number generator.
These forward passes collect attention statistics without optimisation and do not consume the sequence of training batches.
The controller partition of the validation data is used for loss-based acceptance checks and before/after intervention probes; final perplexity uses a separate reporting partition.

\paragraph{Quality and systems measurements.}
The 4K systems sweep holds tokens per update fixed while using micro-batches two, four, and eight with 60, 30, and 15 accumulation steps.
The cross-context systems comparison fixes the number of tokens per micro-batch at 32,768 with $(T,B)=(4096,8),(8192,4),(16384,2)$ and 15 accumulation steps.
Pretraining speedups are medians of within-pair ratios of ordinary-attention to replaced-model median update times; the displayed times are medians across pairs.
These measurements include gradient accumulation, backward computation, clipping, the AdamW update, and gradient reset, but exclude data loading and one-time intervention costs.
Some quality runs used the earlier dense fitter, while subsequent runs used an FFT fitter with separately checked fit agreement. Their intervention times cannot be combined with the update benchmark to establish total training speedup.
Ordinary-attention heads use Flash scaled dot-product attention; every replacement run verifies its head mask, fused dispatch, and zero query/key gradients for replaced heads.
We evaluate on 2,449,408 target tokens from a held-out interval not used to fit the patterns and store per-token log probabilities for paired comparisons.
Using the notation from Section~\ref{sec:experimental-setup}, we report
\begin{equation}
    \deltappl(\%)=100\left(
    \frac{p_{\mathrm{rep}}}{p_{\mathrm{ord}}}-1
    \right).
\end{equation}

Table~\ref{tab:timing-main} compares six one-time replacement stages.
Table~\ref{tab:representation-details} reports the dense and compact pattern representations under both position encodings.
Figure~\ref{fig:prior-quality} gives the fixed-pattern comparison with numerical values in each cell.

\begin{table}[!htbp]
    \centering
    \caption{Timing at 25\% replacement with the post-softmax mean.
    $\tau$: fraction of training completed; $M-t$: remaining updates; $\deltappl$: final perplexity increase.}
    \label{tab:timing-main}
    \begin{tabular}{@{}rrr@{}}
        \toprule
        $\tau$ & $M-t$ & $\deltappl$ (\%) \\
        \midrule
        25\%  & 3750 & 0.724 \\
        40\%  & 3000 & 0.679 \\
        50\%  & 2500 & \textbf{0.662} \\
        60\%  & 2000 & 0.711 \\
        75\%  & 1250 & 0.936 \\
        100\% & 0    & 8.168 \\
        \bottomrule
    \end{tabular}
\end{table}

\begin{table}[!htbp]
    \centering
    \small
    \caption{Single-seed comparison of preset and adaptive replacement schedules.}
    \label{tab:schedule-details}
    \begin{tabular}{llrr}
        \toprule
        Comparison & Setting & Rate & $\deltappl$ (\%) \\
        \midrule
        Preset schedule  & one replacement & 25.00\% & 0.662 \\
                         & gradual, five groups & 25.00\% & 0.689 \\
                         & one replacement & 50.00\% & 2.649 \\
                         & gradual, five groups & 50.00\% & 2.832 \\
        Adaptive schedule & validation budget 1\% & 6.25\% & 0.093 \\
                          & validation budget 2\% & 12.50\% & 0.348 \\
                          & train-loss triggered, $z=3.0$ & 14.58\% & 0.409 \\
                          & train-loss triggered, $z=1.5$ & 50.00\% & 2.119 \\
        \bottomrule
    \end{tabular}
\end{table}

Table~\ref{tab:schedule-confirmation} repeats the one-time and train-loss-triggered schedules with three seeds at matched final rates.
Values are reported as mean $\pm$ standard error, and the time ratio is train-loss-triggered time divided by one-time-schedule time.

\begin{table*}[!htbp]
    \centering
    \small
    \caption{Three-seed comparison of one-time and train-loss-triggered schedules at matched final rates.}
    \label{tab:schedule-confirmation}
    \begin{tabular}{rrrrr}
        \toprule
        Rate & One-time $\deltappl$ (\%) & Train-loss $\deltappl$ (\%) & Difference (pp) & Time ratio (triggered/one-time) \\
        \midrule
        25\% & $0.768\pm0.067$ & $0.858\pm0.180$ & $+0.091\pm0.123$ & 1.027$\times$ \\
        50\% & $2.492\pm0.083$ & $2.495\pm0.189$ & $+0.003\pm0.267$ & 1.066$\times$ \\
        \bottomrule
    \end{tabular}
\end{table*}

Table~\ref{tab:training-systems-details} reports the corresponding post-replacement update speed and peak allocation at the fixed 4K systems configuration.

\begin{table}[!htbp]
    \centering
    \small
    \caption{Post-replacement training speed and memory at $T=4096$ and batch size eight.}
    \label{tab:training-systems-details}
    \begin{tabular}{rrrr}
        \toprule
        Replaced heads & Update speedup & Peak allocated & Saving \\
        \midrule
        0\%  & 1.000$\times$ & 32.21\,GiB & 0.00\,GiB \\
        10\% & 1.016$\times$ & 31.85\,GiB & 0.36\,GiB \\
        25\% & 1.056$\times$ & 31.58\,GiB & 0.63\,GiB \\
        50\% & 1.119$\times$ & 31.24\,GiB & 0.98\,GiB \\
        75\% & 1.215$\times$ & 30.84\,GiB & 1.38\,GiB \\
        \bottomrule
    \end{tabular}
\end{table}

\FloatBarrier
\subsection{Gaussian and Dirichlet distribution diagnostics}
\label{app:dirichlet-fit}

The Gaussian and Dirichlet candidates model variation in different spaces.
The Gaussian candidate fits a normal marginal to each pre-softmax entry across inputs, then applies softmax to obtain a valid probability row.
The Dirichlet candidate instead models the complete post-softmax row jointly: its support already enforces nonnegative entries and a unit row sum.
It therefore needs neither a second softmax nor renormalisation.
In both cases we model variation across calibration inputs, not uncertainty conditioned on the next input.

For the Dirichlet fit, suppress the head indices $(\ell,h)$ and write $\mathbf{m}_i=\widehat{\attn}[i,1{:}i]$ for the empirical mean of causal row $i$.
Let $u_{ij}=(N-1)^{-1}\sum_{n=1}^{N}(\attn^{(n)}(i,j)-\widehat{\attn}(i,j))^2$ be the sample variance of entry $(i,j)$ across the $N$ calibration inputs.
The row model and its moment estimate are
\begin{equation}
    \mathbf{a}_{i}\sim\operatorname{Dirichlet}(c_i\mathbf{m}_i),
    \qquad
    c_i^{\mathrm{mom}}=\frac{1-\lVert\mathbf{m}_i\rVert_2^2}{\sum_{j=1}^{i}u_{ij}}-1,
    \label{eq:dirichlet-moments}
\end{equation}
where $\mathbf{a}_i$ is one sampled row and $c_i$ is its total concentration.
The estimate follows by summing $\operatorname{Var}[a_{ij}]=m_{ij}(1-m_{ij})/(c_i+1)$ over keys, with $m_{ij}$ denoting entry $j$ of $\mathbf{m}_i$.
We multiply the estimated concentration by a chosen scale and bound it to $[10^{-2},10^6]$; nearly deterministic rows use the mean.
For numerical sampling, Dirichlet parameters are floored at $10^{-6}$.
A larger concentration produces rows closer to the mean, but using a single concentration also constrains the covariance between keys.
In particular, for distinct keys $j,k\leq i$, the row model imposes $\operatorname{Cov}[a_{ij},a_{ik}]=-m_{ij}m_{ik}/(c_i+1)$, rather than fitting their covariance independently.
These constraints motivate our evaluation of the fit: valid probability vectors need not reproduce the observed attention distribution.

Our preliminary diagnostics use a 124M checkpoint at update 2,500 with $T=1024$, separate from the main 4K comparison.
We process 500 non-overlapping 1,024-token sequences from the FineWeb-Edu validation split with the model; these are text excerpts not used for pretraining, not model-generated continuations.
For 36 selected heads, we record pre-softmax scores and reconstruct the corresponding probability rows at query positions $i\in\{64,128,256,512,768,1024\}$.
Thus, a fixed head and a fixed pair $(i,j)$ yield 500 scalar observations across different inputs; a fixed head and query yield 500 entire probability vectors.
We fit distributions on 300 sequences and reserve the remaining 200 for visual checks.
Head labels in the diagnostic figures retain the zero-based layer and head indices used in the implementation.

\begin{figure}[!htbp]
    \centering
    \includegraphics[width=0.98\linewidth]{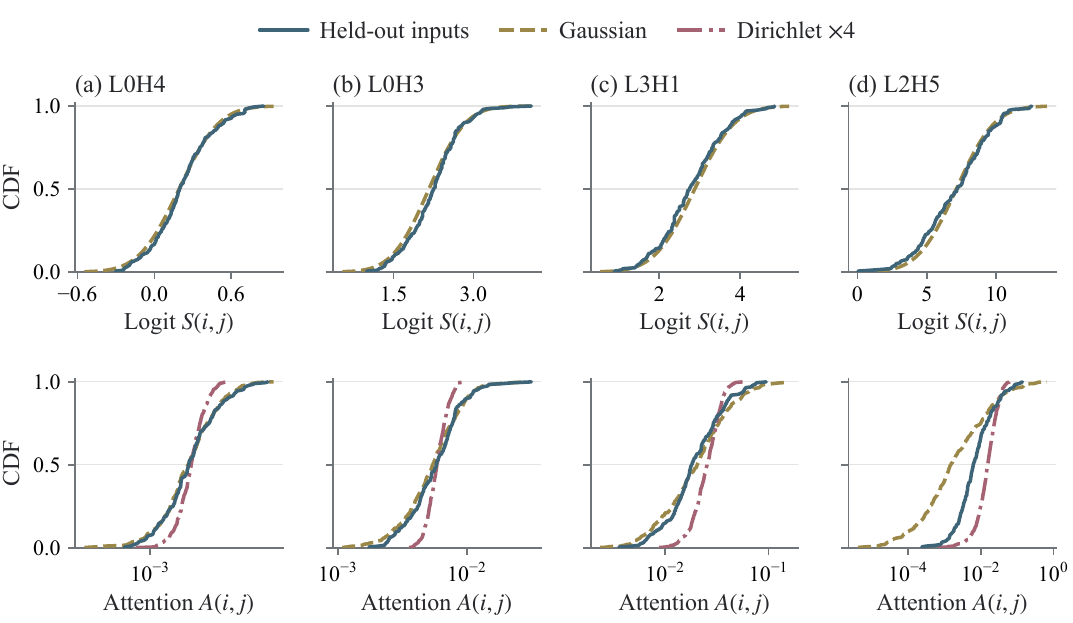}
    \caption{Fixed-entry CDFs at $i=1024$, $j=1023$: logits (top) and attention probabilities (bottom, logarithmic horizontal axis).}
    \label{fig:attention-entry-distributions}
\end{figure}

Figure~\ref{fig:attention-entry-distributions} compares held-out cumulative distributions before and after softmax, using the four heads shown in the original diagnostic.
Each empirical curve uses the 200 held-out sequences; the Gaussian logit curve is analytic, while the probability curves use 500 generated rows per family.
The Gaussian approximates some logit marginals closely, but its fit varies across heads, and sampling logits independently need not preserve the resulting attention distribution.
The Dirichlet samples also differ visibly from the empirical probabilities despite obeying the row-sum constraint.
For a quantitative check beyond individual entries, we compare distributions of complete square-root-transformed rows using energy distance.
Distances are divided by the upper quartile of distances between two empirical subsets over 16 repeated splits, with matched subset sizes of 150 and at most 96 rows per energy estimate.
Across the 36 heads and six query positions, median normalised distances are 1.10 for whole-row resampling from the calibration set, 1.44 for Gaussian sampling, and 2.58 for Dirichlet sampling at concentration scale 4, each averaged over four generated sets before aggregation.
Whole-row resampling provides a reference for finite-sample variability; neither parametric family reproduces every head's distribution.
The concentration scale 4 used in these diagnostics differs from scale 16 in the main pretraining comparison.
The quality of the density fit is distinct from language-model quality when a single sampled pattern is retained.

\begin{figure}[!htbp]
    \centering
    \includegraphics[width=0.98\linewidth]{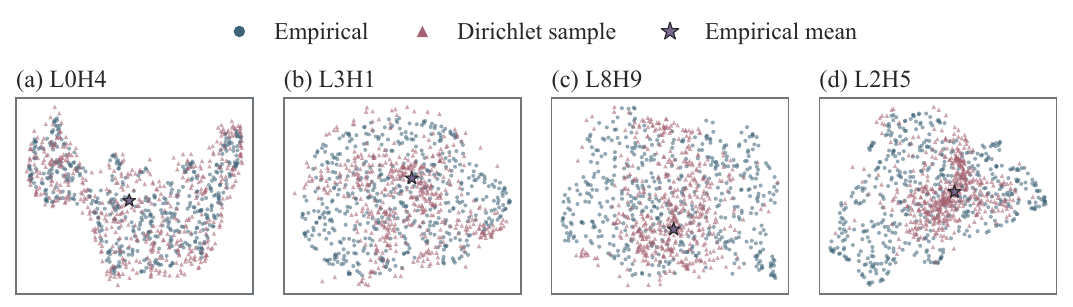}
    \caption{UMAP of empirical binned attention profiles and fitted Dirichlet samples at concentration scale 4.}
    \label{fig:dirichlet-umap}
\end{figure}

Figure~\ref{fig:dirichlet-umap} complements the entry-wise check with a lower-dimensional view of complete attention profiles from the same checkpoint.
This earlier diagnostic uses a separate collection of 128 non-overlapping validation sequences and queries in the later half of each sequence.
For each query row, we sum attention mass into 12 disjoint query--key distance bins: self, previous token, distances 2--3, 4--7, and successive powers-of-two ranges, with the last bin collecting overflow.
Each input therefore gives one 12-dimensional probability vector at each measured query position.
We fit a separate Dirichlet to these binned vectors for each query position and use concentration scale 4.
This is a fit in the binned space, not a fit to individual matrix entries.
For each head, we pool positions and plot 512 empirical and 512 generated profiles.
We apply an elementwise square root followed by coordinate-wise standardisation fitted on the empirical profiles; we then fit UMAP on these empirical profiles and use it to project the generated profiles and the mean.
The square root makes the unstandardised distances proportional to Hellinger distance, but standardisation reweights the coordinates and the UMAP plot is not a quantitative distance test.
Differences in coverage reveal where the fitted samples miss empirical structure; visible clusters alone do not establish a mixture model or identify a uniquely suitable distribution.

\FloatBarrier
\subsection{Replacement and pruning controls}
\label{app:fixed-token-controls}

We compare replacement and pruning first on the same head sets, then with a selector designed for pruning.
The matched-token, matched-time, and later-intervention studies use three pretraining seeds; training the pruned architecture from initialisation uses one.
We then evaluate the resulting checkpoints on downstream tasks and measure the computation retained by fixed-pattern token mixing.

\subsubsection{Matched-token pretraining}

We compare \ourapproach{} with head pruning and random head selection in 124M RoPE models at 4K context.
For each of seeds 1337--1339, we report an ordinary-attention reference and seven continuations, giving 24 complete training paths.
All paths perform 5,000 updates and process 2.4576B target tokens, using micro-batch size eight and 15 accumulation steps on one GH200.
The architecture and AdamW settings follow Appendix~\ref{app:controlled-pretraining-details}: 500 linear warm-up updates to $6\times10^{-4}$, followed by cosine decay to $6\times10^{-5}$ at update 5,000.
These comparisons use shuffled non-overlapping training windows and separate ordinary-attention references from the earlier $B=2$ quality runs.
Each continuation restores the exact parent weights, AdamW state, random-number-generator state, and sampler position; subsequent training data are matched within each seed.
Midpoint replacement occurs at update 2,500; the earlier 25\% setting uses update 1,250.

Calibration uses 32 held-out sequences of length 4,096 in 16 batches of two, disjoint from the reporting interval.
The mean uses 400 Adam fitting steps at learning rate 0.05, without early stopping; these matched-head controls abort if any selected fit exceeds 0.2 nats, rather than substituting another head.
Across the twelve midpoint mean runs (three seeds, two rates, and two budgets), the selected sets equal the raw variance top-$k$ sets: no heads are rejected or substituted, and the largest selected-head fitting KL is 0.159 nats.
Thus, the fitting constraint does not change which heads are compared in these controls; this check concerns the selected sets, not all 144 candidates.
Pruning uses exactly the same heads as the variance-selected mean at each rate.
Pruning removes the selected heads' contributions, their query/key/value projection rows, and the corresponding output-projection columns, preserving the optimiser state of retained parameters.
Random selection draws heads within each layer, matching the variance-selected counts and applying the same fitting threshold; it retains fitted mean patterns rather than using random patterns.
The earlier intervention recalibrates and selects heads at its own checkpoint.

All final perplexities use the same 2,449,408 reporting tokens, with attention-projection caches refreshed before evaluation.
Relative changes are computed against each seed's ordinary-attention reference before averaging; standard errors use the sample standard deviation divided by $\sqrt{3}$.
Table~\ref{tab:fixed-token-controls-full} combines absolute results with paired differences from midpoint mean replacement at the same rate.
Relative perplexity changes appear in Table~\ref{tab:fixed-token-controls-main}; the additional early 25\% setting gives $0.772\pm0.021\%$.
Total training time includes training with ordinary attention before replacement and continuation afterwards, including calibration, head scoring, fitting, compilation, periodic monitoring, and intermediate checkpointing.
It excludes queueing, final reporting evaluation, and checkpoint export.
Configurations run in separate GH200 allocations; these full-path times differ from the controlled paired update benchmarks in Table~\ref{tab:main-results}.
The matched-time comparison in Appendix~\ref{app:equal-time-controls} uses an elapsed-time learning-rate schedule instead.

\begin{table}[!htbp]
    \centering
    \footnotesize
    \setlength{\tabcolsep}{3pt}
    \caption{Matched-token controls: mean $\pm$ SE over three pretraining seeds. Paired columns subtract midpoint mean + variance at the same rate.}
    \label{tab:fixed-token-controls-full}
    \begin{tabular}{@{}lrrrrr@{}}
        \toprule
        & & & & \multicolumn{2}{c}{Paired differences} \\
        \cmidrule(l){5-6}
        Setting (rate) & PPL $\downarrow$ & \shortstack{Training\\(min)} & \shortstack{Intervention\\(s)} & $\deltappl$ (pp) & Time (s) \\
        \midrule
        Ordinary attention & $23.896\pm0.038$ & $188.7\pm2.7$ & --- & --- & --- \\
        Mean + variance (25\%) & $24.054\pm0.041$ & $184.7\pm1.6$ & $54.9\pm6.0$ & --- & --- \\
        Mean + variance (50\%) & $24.492\pm0.035$ & $178.7\pm2.0$ & $46.3\pm1.2$ & --- & --- \\
        Pruning + variance (25\%) & $24.088\pm0.042$ & $178.9\pm1.6$ & $46.3\pm3.1$ & $+0.1428\pm0.0040$ & $-343.9\pm27.8$ \\
        Pruning + variance (50\%) & $24.621\pm0.049$ & $168.3\pm1.3$ & $50.8\pm5.8$ & $+0.5378\pm0.0622$ & $-620.9\pm65.3$ \\
        Mean + random (25\%) & $24.202\pm0.033$ & $183.4\pm1.2$ & $53.7\pm5.6$ & $+0.6186\pm0.0743$ & $-74.2\pm23.7$ \\
        Mean + random (50\%) & $24.763\pm0.026$ & $178.8\pm1.8$ & $57.5\pm2.8$ & $+1.1344\pm0.0695$ & $+9.1\pm63.6$ \\
        \shortstack[l]{Mean + variance\\(25\%, early)} & $24.080\pm0.037$ & $183.0\pm1.1$ & $49.2\pm4.1$ & $+0.1085\pm0.0170$ & $-99.7\pm31.0$ \\
        \bottomrule
    \end{tabular}
\end{table}

\paragraph{One-time intervention cost.}
Table~\ref{tab:intervention-costs} separates attention measurement from fitting the compact parameters.
For the 124M controls, total intervention time includes calibration, ranking, fitting, the KL check, and installation.
Fitting and installation together take $3.30\pm0.17$\,s at 25\% replacement and $5.40\pm0.10$\,s at 50\%; ranking takes less than 0.012\,s in every run.
The two rates have separate calibration measurements, which account for the variation in total intervention time.
For the 1B runs, the total is measured on rank zero through synchronisation and broadcast of the fitted state; ranking takes 0.019 and 0.021\,s at 25\% and 50\% replacement, respectively.
Separate before/after diagnostic evaluations add 3.68 and 2.76\,s at 1B and are excluded from this table.
The 124M entries apply to these $B=8$ controls using the FFT fitter; the earlier $B=2$ quality experiments include both dense and FFT fitting implementations.

\begin{table}[!htbp]
    \centering
    \small
    \caption{One-time calibration and fitting costs in seconds. The 124M controls report mean $\pm$ SE over three seeds; the 1B runs use one seed. Fitting is included in the total.}
    \label{tab:intervention-costs}
    \begin{tabular}{@{}lcrrr@{}}
        \toprule
        Model & Rate & Capture and mean & Fit $\alpha,\rho$ & Total intervention \\
        \midrule
        124M, 4K & 25\% & $51.56\pm5.96$ & $3.20\pm0.16$ & $54.86\pm6.04$ \\
        124M, 4K & 50\% & $40.89\pm1.11$ & $5.24\pm0.09$ & $46.30\pm1.22$ \\
        1B, 8K & 25\% & 1422.37 & 39.56 & 1464.41 \\
        1B, 8K & 50\% & 1313.35 & 76.28 & 1392.64 \\
        \bottomrule
    \end{tabular}
\end{table}

\subsubsection{Pattern content and trainability}
\label{app:pattern-controls}

We compare frozen and trainable compact means at 25\% replacement in 124M RoPE models at 4K context, with $B=8$ and 15 accumulation steps.
Across three pretraining seeds, both variants restore the same 2,500-update ordinary-attention checkpoint, AdamW state, and training-data order, then continue to 5,000 updates and 2.4576B tokens.
They use the same 36 variance-selected heads and exactly the same fitted initial parameters, with every selected fit below the 0.2-nat threshold.
Only the trainable variant adds $\alpha,\rho$ to AdamW, using the model learning rate and betas, zero weight decay, and initially zero optimiser moments for these new parameters.
Both variants use the fused mixed-head forward and value-gradient implementation; the trainable variant additionally computes pattern gradients and refreshes the normalisers after each update.

Table~\ref{tab:pattern-controls}a reports fresh frozen and trainable runs in separate GH200 allocations.
Update time averages the last 100 complete optimiser updates; continuation time includes loading, intervention, compilation, and monitoring, with final reporting and checkpoint export excluded.
Perplexity uses the same reporting tokens as Table~\ref{tab:fixed-token-controls-full}.
The paired relative PPL change from learning the patterns is $-0.010\pm0.006\%$, while update and continuation times increase by $11.9\pm1.2\%$ and $12.4\pm1.5\%$, respectively.

For MQAR, we use the ordinary-attention and frozen-mean checkpoints from Table~\ref{tab:fixed-token-controls-full}, together with the trainable endpoints from Table~\ref{tab:pattern-controls}a.
All three groups use matched 2.4576B-token pretraining budgets, with midpoint replacement where applicable.
The task protocol follows Appendix~\ref{app:associative-recall}, with task seed 1337 for each pretraining seed: 1,500 updates at 512 tokens and eight pairs.
Trainable patterns continue learning during task adaptation with zero pattern weight decay; the remaining model parameters use weight decay 0.01.
Training data and all 512-example test sets are shared across methods and pretraining seeds.
This comparison evaluates pattern choices at the midpoint; Figure~\ref{fig:mqar-generalisation} evaluates pruning after later, matched-time pretraining.

In Table~\ref{tab:pattern-controls}b, continued pattern learning gives slightly higher mean accuracy than frozen mean in the length-only tests, whereas frozen mean gives higher means throughout the fixed-length pair-count sweep.

\begin{table}[!htbp]
    \centering
    \footnotesize
    \setlength{\tabcolsep}{4pt}
    \caption{Pattern controls at 25\% replacement: mean $\pm$ SE over three pretraining seeds.}
    \label{tab:pattern-controls}
    \begin{tabular}{@{}lrrrr@{}}
        \toprule
        \multicolumn{5}{@{}l}{\emph{(a) Matched-token pretraining: 124M, 4K, 2.4576B tokens}} \\
        Pattern & PPL $\downarrow$ & Update (s) $\downarrow$ & Continuation (min) $\downarrow$ & Peak (GiB) $\downarrow$ \\
        \midrule
        Frozen mean & $24.055\pm0.040$ & $2.159\pm0.012$ & $90.96\pm0.56$ & $31.668\pm0.002$ \\
        Trainable mean & $24.052\pm0.042$ & $2.416\pm0.013$ & $102.27\pm0.82$ & $31.801\pm0.002$ \\
        \bottomrule
    \end{tabular}
    \par\medskip
    \begin{tabular}{@{}rrrrr@{}}
        \toprule
        \multicolumn{5}{@{}l}{\emph{(b) MQAR accuracy (\%) $\uparrow$: midpoint, matched-token checkpoints}} \\
        Tokens & Pairs & Ordinary attn. & Frozen mean & Trainable mean \\
        \midrule
        512 & 8  & $99.58\pm0.15$ & $99.60\pm0.06$ & $99.54\pm0.07$ \\
        512 & 16 & $71.78\pm2.47$ & $81.36\pm1.58$ & $78.44\pm3.69$ \\
        512 & 24 & $52.71\pm2.06$ & $60.89\pm1.61$ & $59.67\pm5.69$ \\
        512 & 32 & $42.13\pm1.49$ & $49.08\pm3.04$ & $47.77\pm5.62$ \\
        512 & 48 & $30.00\pm1.30$ & $35.39\pm3.74$ & $34.67\pm5.00$ \\
        512 & 64 & $23.53\pm1.34$ & $28.30\pm4.18$ & $27.04\pm4.19$ \\
        \addlinespace[2pt]
        2,048 & 8 & $97.01\pm0.46$ & $97.53\pm0.44$ & $97.99\pm0.73$ \\
        4,096 & 8 & $88.76\pm0.88$ & $88.89\pm0.93$ & $89.65\pm1.42$ \\
        \bottomrule
    \end{tabular}
\end{table}

\subsubsection{Matched-time pretraining}
\label{app:equal-time-controls}

To test whether cheaper updates improve quality within a fixed training time, we restore each seed's common 2,500-update checkpoint.
The remaining allowance is the ordinary-attention reference's total training time minus the time spent reaching the shared checkpoint.
Every continuation receives this remaining allowance, including loading, compilation, calibration, selection, fitting, installation, and monitoring.
Queueing and final reporting or checkpoint export are excluded.
The learning-rate schedule advances from its midpoint value to $6\times10^{-5}$ as a function of elapsed time, identically for ordinary attention, mean, pruned, and random-head models.
Training follows the same data order within each seed, but faster models process more tokens.
All final perplexities use the 2,449,408-token reporting set from the matched-token comparison.

Table~\ref{tab:equal-time-controls} gives absolute perplexities and update counts for the comparison in Table~\ref{tab:fixed-token-controls-main}.
The full-path budgets span 187.3--188.1 minutes across seeds; differences of a few seconds within a seed arise from stopping before the next update would exceed the allowance.
Runs use separate GH200 allocations; the paired kernel benchmarks instead alternate models on the same GPU.
Random selection preserves the variance-selected number of heads in each layer and uses fitted mean patterns under the same time allowance.
Its perplexity penalty exceeds variance selection by $0.786\pm0.082$ and $0.866\pm0.154$ percentage points at 25\% and 50\%, respectively, with the same ordering in all three seeds.

\begin{table}[!htbp]
    \centering
    \footnotesize
    \setlength{\tabcolsep}{4pt}
    \caption{Matched-time controls: mean $\pm$ SE over three pretraining seeds.
    Updates and times include the initial training with ordinary attention.}
    \label{tab:equal-time-controls}
    \begin{tabular}{@{}lcrrr@{}}
        \toprule
        Operation + selection & Rate & PPL $\downarrow$ & Updates & Time (min) \\
        \midrule
        Ordinary attention & 0\% & $23.882\pm0.023$ & $5008\pm13$ & $187.65\pm0.25$ \\
        Mean + variance & 25\% & $23.991\pm0.006$ & $5109\pm11$ & $187.64\pm0.25$ \\
        Mean + variance & 50\% & $24.338\pm0.020$ & $5221\pm18$ & $187.64\pm0.25$ \\
        Pruning + variance & 25\% & $23.909\pm0.024$ & $5267\pm30$ & $187.64\pm0.25$ \\
        Pruning + variance & 50\% & $24.115\pm0.003$ & $5685\pm36$ & $187.65\pm0.25$ \\
        Mean + random heads & 25\% & $24.178\pm0.024$ & $5097\pm28$ & $187.64\pm0.25$ \\
        Mean + random heads & 50\% & $24.545\pm0.048$ & $5242\pm11$ & $187.65\pm0.26$ \\
        \bottomrule
    \end{tabular}
\end{table}

\subsubsection{Later replacement and pruning-specific selection}
\label{app:pruning-selection-budget}

We compare 25\% replacement and pruning across three pretraining seeds after 5,000 updates with ordinary attention, or 2.4576B tokens.
Within each seed, all continuations restore the same checkpoint and optimiser state, retaining 4K context, micro-batch eight, and 15 accumulation steps.
The token-matched group adds 2,500 updates, giving 3.6864B tokens in total.
Its continuation learning rate decays from $6\times10^{-5}$ to $6\times10^{-6}$.
The time-matched group instead receives its seed's measured token-matched continuation time with ordinary attention, 5,648--5,670 seconds, and uses the same endpoint learning rates with elapsed-time cosine decay.
The allowance includes selection and installation; continuations stop before another update would exceed it.
Final reporting uses 4,997,120 held-out tokens, so absolute perplexities should be compared within this study rather than with Table~\ref{tab:equal-time-controls}.

The pruning-specific criterion measures loss sensitivity through head-output gates \citep{NEURIPS2019_2c601ad9}.
For calibration sequence $n$, multiply head $\ell h$'s output by a scalar gate $\xi_{\ell h}$ before the output projection, and let $\mathcal{L}^{(n)}(\boldsymbol{\xi})$ be its mean-token negative log-likelihood with gate collection $\boldsymbol{\xi}$.
Ordinary attention has all gates equal to one.
The gate-Taylor importance and normalised pruning score are
\begin{equation}
    I_{\ell h}=\frac{1}{N}\sum_{n=1}^{N}
    \left|\left.\frac{\partial\mathcal{L}^{(n)}(\boldsymbol{\xi})}
    {\partial\xi_{\ell h}}\right|_{\boldsymbol{\xi}=\mathbf{1}}\right|,
    \qquad
    s_{\ell h}^{\mathrm{gate}}
    =\frac{I_{\ell h}}{\sqrt{\sum_{h'=1}^{H}I_{\ell h'}^2}}.
    \label{eq:gate-taylor-pruning}
\end{equation}
Zero-norm layers receive zero scores.
We take the absolute derivative before averaging across sequences and remove the 36 heads with the smallest normalised scores globally, breaking ties by layer and head index.
This is a one-time use of the importance criterion, not the iterative pruning procedure of \citet{NEURIPS2019_2c601ad9}.
Calibration uses exactly the same 32 held-out sequences as variance selection.
In the matched-time group, the selected set overlaps with the variance-selected set in 19, 19, and 21 of 36 heads across seeds; per-layer removal counts can also differ.
Both pruning variants remove the corresponding projection rows and columns and preserve the retained AdamW state.
\ourapproach{} and variance pruning retain exactly the same selected heads across token and time budgets.
Independent gate-Taylor scoring selects 36, 35, and 35 common heads across budgets: two seeds exchange the nearly tied 36th and 37th candidates.
The gate-Taylor budget comparison therefore also includes this small selection difference.

Table~\ref{tab:pruning-additional-training} combines pretraining quality with finetuning of the matched-time checkpoints.
The paired finetuning differences, \ourapproach{} minus gate-Taylor, are $+0.765\pm0.894$ points on SST-2 and $-0.051\pm0.795$ on BoolQ, with reversals across pretraining seeds.

Training the reference-derived, variance-pruned architecture from initialisation for 7,500 updates gives PPL 22.683 in seed 1337, compared with 23.049 when introducing the same pruning after 5,000 updates.
Its 248.26-minute training time excludes the cost of obtaining the reference-derived layout; the corresponding full-path times for ordinary attention, \ourapproach{}, variance pruning, and gate-Taylor pruning are 286.86, 282.87, 275.43, and 275.93 minutes in this seed.

\begin{table}[!htbp]
    \centering
    \footnotesize
    \setlength{\tabcolsep}{3pt}
    \caption{Later 25\% intervention: mean $\pm$ SE across three pretraining seeds. Finetuning uses the matched-time checkpoints and one fixed seed.}
    \label{tab:pruning-additional-training}
    \begin{tabular}{@{}lrrrr@{}}
        \toprule
        & \multicolumn{2}{c}{Pretraining PPL $\downarrow$} & \multicolumn{2}{c}{Finetuning accuracy (\%) $\uparrow$} \\
        \cmidrule(lr){2-3}\cmidrule(l){4-5}
        Operation + selection & Matched tokens & Matched time & SST-2 & BoolQ \\
        \midrule
        Ordinary attention & $22.563\pm0.010$ & $22.561\pm0.011$ & $87.00\pm0.33$ & $71.78\pm0.46$ \\
        \ourapproach{} + variance & $22.923\pm0.013$ & $22.895\pm0.012$ & $87.73\pm0.46$ & $71.27\pm0.67$ \\
        Pruning + variance & $23.019\pm0.028$ & $22.953\pm0.026$ & $87.81\pm0.34$ & $71.75\pm0.29$ \\
        Pruning + gate-Taylor & $22.865\pm0.019$ & $22.798\pm0.021$ & $86.96\pm0.49$ & $71.33\pm0.16$ \\
        \bottomrule
    \end{tabular}
\end{table}

\subsubsection{Downstream finetuning}
\label{app:pruning-downstream-systems}

Tables~\ref{tab:pruning-additional-training} and~\ref{tab:pruning-downstream} distinguish pretraining variation from finetuning variation.
The former and the midpoint group in the latter vary pretraining seeds while fixing the finetuning seed at 1337.
The supplementary table also retains the single-seed token-matched controls and repeats finetuning with three seeds on the seed-1337 matched-time checkpoints.
SST-2 and BoolQ use batch size 16, dynamic padding, and length caps of 128 and 512 tokens, respectively.
The learning rate is $2\times10^{-5}$ and AdamW weight decay is 0.01.
Within each pretraining group and task, the ordinary-attention model selects an epoch from one to five on a stratified 10\% internal development split; all alternatives use that epoch count and the same training examples in the same order.
The official validation set is evaluated after selection.
Downstream updates are matched within each seed and task, not combined pretraining and finetuning time.

The pretraining-seed and finetuning-seed comparisons share seed 1337 and are not pooled as independent repetitions.

\begin{table}[!htbp]
    \centering
    \small
    \setlength{\tabcolsep}{3pt}
    \caption{Supplementary finetuning comparisons. Each group specifies which seed varies; errors are standard errors.}
    \label{tab:pruning-downstream}
    \begin{tabular}{@{}lcrr@{}}
        \toprule
        & & \multicolumn{2}{c}{Accuracy (\%) $\uparrow$} \\
        \cmidrule(l){3-4}
        Operation + selection & Rate & SST-2 & BoolQ \\
        \midrule
        \multicolumn{4}{@{}l}{\emph{Midpoint, matched time: three pretraining seeds, one finetuning seed}} \\
        Ordinary attention & 0\% & $86.24\pm0.93$ & $71.86\pm0.40$ \\
        Mean + variance & 25\% & $86.05\pm0.54$ & $70.52\pm0.16$ \\
        Mean + variance & 50\% & $86.12\pm0.54$ & $70.66\pm0.26$ \\
        Pruning + variance & 25\% & $86.85\pm0.56$ & $70.97\pm0.66$ \\
        Pruning + variance & 50\% & $85.93\pm0.56$ & $69.92\pm0.26$ \\
        \midrule
        \multicolumn{4}{@{}l}{\emph{Later, matched tokens: one pretraining and one finetuning seed}} \\
        Ordinary attention & 0\% & 88.88 & 71.74 \\
        Mean + variance & 25\% & 87.39 & 72.11 \\
        Pruning + variance & 25\% & 87.73 & 71.99 \\
        Pruning + gate-Taylor & 25\% & 87.61 & 71.87 \\
        Pruning from initialisation & 25\% & 89.11 & 71.25 \\
        \midrule
        \multicolumn{4}{@{}l}{\emph{Later, matched time: one pretraining seed, three finetuning seeds}} \\
        Ordinary attention & 0\% & $87.81\pm0.28$ & $71.54\pm0.23$ \\
        Mean + variance & 25\% & $87.69\pm0.44$ & $71.78\pm0.48$ \\
        Pruning + variance & 25\% & $87.31\pm0.37$ & $71.49\pm0.31$ \\
        Pruning + gate-Taylor & 25\% & $87.65\pm0.23$ & $71.20\pm0.03$ \\
        \bottomrule
    \end{tabular}
\end{table}

\subsubsection{Associative recall: adaptation and immediate replacement}
\label{app:associative-recall}

Multi-query associative recall requires predicting the value associated with a key that appeared earlier in the sequence \citep{ICLR2024_448fc91f}.
We use the unmodified Zoology generator at revision \texttt{1ad20d1}, with symbolic vocabulary size 8,192, random non-query tokens, and \texttt{power\_a}=0.01 for the distance distribution.
The four matched-time models from Table~\ref{tab:pruning-additional-training} retain their fitted patterns or pruned layouts throughout adaptation; no heads are selected again.
Each model receives 1,500 updates on 512-token sequences with eight key-value pairs, using batch size 16 and AdamW with learning rate $10^{-4}$ and weight decay 0.01.
Loss and accuracy are measured only at supervised query positions, with all 50,304 model output classes available.

We vary three pretraining seeds with task seed 1337 fixed, then vary three task-training seeds with pretraining seed 1337 fixed.
Both groups share the same seed-1337 experiment and are not pooled as six independent repetitions.
The training stream is shared across methods within each task seed, with development and test data fixed across both seed axes.
The ordinary-attention model must reach 80\% accuracy on 256 development examples before comparing alternatives; all runs pass this predeclared check.
Evaluation uses the final update rather than a test-selected checkpoint, with 512 examples per condition.
The three-pretraining-seed comparison tests 8, 16, 24, 32, 48, and 64 pairs at 512 tokens, without further training, fitting, or head selection (Figure~\ref{fig:mqar-generalisation}).
All methods and seeds share evaluation examples, generated with seed $800001+T+k$, where $T$ is sequence length and $k$ is pair count; evaluation uses micro-batch size four.
Increasing the pair count also changes query placement and density under the unchanged generator.
Table~\ref{tab:mqar-complete} combines this curve with the length-only tests, task-seed replication, and immediate-replacement study below.

The accuracy gap widens as more associations must be retrieved: the mean advantage over variance pruning grows from 5.65 points at 16 pairs to 25.39 points at 64 pairs.
\ourapproach{} exceeds both pruning variants in every pretraining seed at 24--64 pairs; at 16 pairs, seed 1338 does not share the mean advantage.
The separate task-seed experiment confirms the ordering at 32 pairs, but does not evaluate the additional pair counts.
The length-only tests instead favour pruning, distinguishing generalisation to more associations from generalisation to longer inputs.

To distinguish adaptation from immediate retention, we also apply all interventions to one ordinary-attention model already trained on MQAR, using the seed-1337 task checkpoint above.
Calibration uses 32 held-out task sequences at the native 4K context: variance selection with the 0.2-nat compact-fitting threshold selects 36 heads, and variance pruning removes exactly those heads.
Gate-Taylor selection uses the query-only loss in Equation~\ref{eq:gate-taylor-pruning}.
Every arm, including ordinary attention, starts fresh AdamW state because the task checkpoint contains no optimiser state, then receives the same new 512-token, eight-pair training stream at the adaptation learning rate and batch size.
Table~\ref{tab:mqar-complete} includes all four test conditions immediately after intervention and after 50, 200, and 500 recovery updates.
Immediate replacement does not reproduce the consistent 32-pair advantage of the fully adapted \ourapproach{} models; the subsequent ordering also changes during recovery.
The two studies differ in both intervention stage and calibration domain, so they distinguish the measured behaviours without isolating a single cause.

\begin{table}[ht]
    \centering
    \footnotesize
    \setlength{\tabcolsep}{5pt}
    \caption{MQAR query accuracy (\%).
    Adaptation varies one seed axis at a time (mean $\pm$ SE); immediate replacement and recovery use one task-trained ordinary-attention parent.}
    \label{tab:mqar-complete}
    \begin{tabular}{@{}llrrrr@{}}
        \toprule
        & & & & \multicolumn{2}{c}{Pruning 25\%} \\
        \cmidrule(l){5-6}
        Updates & Tokens / pairs & Ordinary attn. & \ourapproach{} 25\% & Variance & Gate-Taylor \\
        \midrule
        \multicolumn{6}{@{}l}{\emph{Adaptation: three pretraining seeds, task seed 1337}} \\
        1,500 & 512 / 8 & $99.58\pm0.05$ & $99.73\pm0.04$ & $99.72\pm0.03$ & $99.62\pm0.06$ \\
         & 512 / 16 & $81.32\pm3.21$ & $91.08\pm1.58$ & $85.43\pm3.07$ & $80.79\pm4.50$ \\
         & 512 / 24 & $60.56\pm4.22$ & $79.76\pm2.72$ & $66.25\pm3.22$ & $60.59\pm6.42$ \\
         & 512 / 32 & $47.54\pm3.87$ & $70.83\pm4.86$ & $52.32\pm2.41$ & $47.95\pm6.06$ \\
         & 512 / 48 & $34.01\pm3.38$ & $61.14\pm7.37$ & $37.70\pm1.54$ & $34.49\pm5.42$ \\
         & 512 / 64 & $26.19\pm2.75$ & $54.42\pm8.70$ & $29.03\pm1.42$ & $26.59\pm4.41$ \\
         & 2,048 / 8 & $98.27\pm0.30$ & $97.96\pm0.55$ & $98.66\pm0.11$ & $98.65\pm0.26$ \\
         & 4,096 / 8 & $90.35\pm0.58$ & $89.67\pm1.09$ & $90.74\pm0.36$ & $91.43\pm0.60$ \\
        \midrule
        \multicolumn{6}{@{}l}{\emph{Adaptation: pretraining seed 1337, three task-training seeds}} \\
        1,500 & 512 / 8 & $99.71\pm0.09$ & $99.76\pm0.05$ & $99.75\pm0.05$ & $99.72\pm0.07$ \\
         & 2,048 / 8 & $98.01\pm0.28$ & $97.58\pm0.01$ & $98.87\pm0.21$ & $98.78\pm0.19$ \\
         & 4,096 / 8 & $89.67\pm0.32$ & $88.21\pm0.28$ & $91.06\pm0.46$ & $90.76\pm0.27$ \\
         & 512 / 32 & $44.55\pm1.73$ & $73.20\pm4.26$ & $52.13\pm5.56$ & $53.20\pm5.25$ \\
        \midrule
        \multicolumn{6}{@{}l}{\emph{Immediate replacement (0) and subsequent recovery updates}} \\
        0 & 512 / 8 & 99.54 & 99.68 & 99.68 & 99.66 \\
         & 2,048 / 8 & 97.75 & 97.00 & 97.71 & 97.19 \\
         & 4,096 / 8 & 89.21 & 88.75 & 89.48 & 88.79 \\
         & 512 / 32 & 47.05 & 48.75 & 60.83 & 52.58 \\
        \addlinespace[2pt]
        50 & 512 / 8 & 99.61 & 99.66 & 99.78 & 99.37 \\
         & 2,048 / 8 & 96.88 & 98.27 & 97.61 & 95.61 \\
         & 4,096 / 8 & 87.65 & 90.28 & 89.38 & 86.57 \\
         & 512 / 32 & 51.93 & 43.45 & 51.54 & 53.65 \\
        \addlinespace[2pt]
        200 & 512 / 8 & 99.68 & 99.46 & 99.83 & 99.17 \\
         & 2,048 / 8 & 97.46 & 96.14 & 98.61 & 96.61 \\
         & 4,096 / 8 & 88.79 & 87.11 & 90.48 & 87.96 \\
         & 512 / 32 & 61.99 & 52.70 & 56.05 & 46.50 \\
        \addlinespace[2pt]
        500 & 512 / 8 & 99.73 & 99.73 & 99.85 & 99.76 \\
         & 2,048 / 8 & 97.73 & 97.51 & 97.90 & 97.09 \\
         & 4,096 / 8 & 89.16 & 88.28 & 89.45 & 88.33 \\
         & 512 / 32 & 44.74 & 58.39 & 56.79 & 48.14 \\
        \bottomrule
    \end{tabular}
\end{table}

\subsubsection{Training and prefill costs}
\label{app:pruning-execution-costs}

Pruning removes more computation than fixed-pattern replacement because the selected heads no longer compute values or their output projections.
Table~\ref{tab:pruning-native-systems} measures this difference at 4K, 8K, and 16K using the largest tested batch accepted by all compared methods under a 90\% peak-allocation limit.
Each update accumulates six micro-batches, totalling 491,520 tokens, and includes the forward pass, full-vocabulary loss, backward pass, clipping, and AdamW.
Data loading, fitting, compilation, and evaluation are excluded.
Pruned models are constructed from the ordinary-attention native-context checkpoints using the replaced models' head sets, solely for these systems measurements.
Both methods reduce update time at each reported context, with pruning giving larger savings.
The 8K and 16K pruned models are used only for timing; this comparison does not include their quality after adaptation.

\begin{table}[!htbp]
    \centering
    \footnotesize
    \setlength{\tabcolsep}{3pt}
    \caption{Native-context training-update costs for matched head sets.
    Memory changes are relative to ordinary attention at the same context and batch size.}
    \label{tab:pruning-native-systems}
    \begin{tabular}{@{}lcrrrrrr@{}}
        \toprule
        & & & \multicolumn{3}{c}{Update time (ms) $\downarrow$} & \multicolumn{2}{c}{Peak change (\%)} \\
        \cmidrule(lr){4-6}\cmidrule(l){7-8}
        Context & $B$ & Rate & Ordinary attn. & Mean & Pruning & Mean & Pruning \\
        \midrule
        4K & 20 & 25\% & 2172.80 & 2057.35 & 1923.26 & $-2.15$ & $-2.19$ \\
        4K & 20 & 50\% & 2173.12 & 1929.10 & 1678.24 & $-3.34$ & $-4.22$ \\
        8K & 10 & 25\% & 2651.80 & 2573.68 & 2307.18 & $-2.24$ & $-2.88$ \\
        8K & 10 & 50\% & 2652.83 & 2408.64 & 1933.85 & $-3.27$ & $-4.95$ \\
        16K & 5 & 25\% & 3714.63 & 3621.11 & 3083.03 & $-1.76$ & $-5.08$ \\
        16K & 5 & 50\% & 3716.74 & 3419.69 & 2472.20 & $-2.82$ & $-7.64$ \\
        \bottomrule
    \end{tabular}
\end{table}

Finally, Table~\ref{tab:pruning-prefill} measures full-model causal prefill on the additional-training checkpoints before task finetuning.
The benchmark includes the final-token vocabulary projection and uses FP32 stored weights and prior buffers with BF16 autocast.
Each alternative has four paired ordinary-attention measurements with alternating execution order, fresh model processes, ten warm-up calls, and forty timed calls.
Latencies are mean synchronised wall times; speedups are mean paired ratios with SE over timing repeats.
These prefill measurements use different workloads and precision settings from Table~\ref{tab:pruning-native-systems}; their memory changes should not be interchanged.
At $B=1$, mean replacement is slower, with speedups of 0.933$\times$ and 0.877$\times$ for the token- and time-matched groups, respectively.
All measurements concern causal prefill without token-by-token KV-cache decoding.

\begin{table}[!htbp]
    \centering
    \footnotesize
    \setlength{\tabcolsep}{3pt}
    \caption{Causal prefill at $B=32$, $T=4096$ for the single-seed 25\% controls. Speedup SE is over four timing pairs.}
    \label{tab:pruning-prefill}
    \begin{tabular}{@{}lrrrr@{}}
        \toprule
        Operation + selection & Ordinary attn. (ms) & Method (ms) & Speedup ($\times$) & Peak change (\%) \\
        \midrule
        \multicolumn{5}{@{}l}{\emph{Matched-token pretraining}} \\
        Mean + variance & 154.59 & 142.52 & $1.0847\pm0.0024$ & $+3.75$ \\
        Pruning + variance & 155.26 & 133.43 & $1.1636\pm0.0012$ & $-4.75$ \\
        Pruning + gate-Taylor & 154.22 & 132.63 & $1.1628\pm0.0009$ & $-5.66$ \\
        Pruning from initialisation & 154.27 & 132.64 & $1.1631\pm0.0005$ & $-4.75$ \\
        \midrule
        \multicolumn{5}{@{}l}{\emph{Matched-time pretraining}} \\
        Mean + variance & 153.29 & 141.34 & $1.0846\pm0.0013$ & $+3.75$ \\
        Pruning + variance & 153.55 & 131.91 & $1.1640\pm0.0007$ & $-4.75$ \\
        Pruning + gate-Taylor & 153.54 & 131.72 & $1.1657\pm0.0011$ & $-5.66$ \\
        \bottomrule
    \end{tabular}
\end{table}

\FloatBarrier
\subsection{Finetuning the 124M checkpoints}
\label{app:downstream-details}

For the 124M models, accuracy values are reported as mean $\pm$ standard error over three finetuning seeds from one pretrained checkpoint.
Timing and memory measurements use three batch-order seeds at the task-specific micro-batch sizes shown below; QuALITY uses four accumulation steps for an effective batch of 16.
Each timed update includes all accumulation steps, the task loss, backward pass, gradient clipping, AdamW update, and gradient reset.
For paired ordinary-attention and replaced-model update times $t_{\mathrm{ord},r}$ and $t_{\mathrm{rep},r}$, finetuning speedup is $\bar{s}=\frac{1}{3}\sum_{r=1}^{3}t_{\mathrm{ord},r}/t_{\mathrm{rep},r}$; the displayed times are averaged separately.
For example, the SST-2 25\% speedups are 0.9633, 0.6653, and 1.0732, yielding $0.901\pm0.122$, whereas the ratio of mean times is 0.861.
The corresponding replaced update times span 46.23--77.15\,ms across the three timing seeds; the table retains this variation through the reported standard errors.
Table~\ref{tab:downstream-details} adds uncertainty estimates to the summary in Table~\ref{tab:main-results}; Figure~\ref{fig:systems-across-stages}b shows the paired finetuning speedups.

\begin{table*}[!htbp]
    \centering
    \scriptsize
    \setlength{\tabcolsep}{3.5pt}
    \caption{Finetuning quality, update time, and peak-memory change for inherited fixed-attention checkpoints.}
    \label{tab:downstream-details}
    \begin{tabular}{@{}lrrrrrr@{}}
        \toprule
        Task / checkpoint & $B$ & Accuracy (\%) & $\Delta$ accuracy (pp) & Time/update & Speedup & Peak memory change \\
        \midrule
        SST-2, ordinary attn. & 256 & $86.39\pm0.27$ & --- & $50.11\pm0.61$\,ms & 1.000$\times$ & 0.00\% \\
        SST-2, fixed 25\% & 256 & $86.54\pm0.36$ & $+0.15\pm0.53$ & $58.22\pm9.58$\,ms & $0.901\pm0.122\times$ & $-3.65\%$ \\
        SST-2, fixed 50\% & 256 & $86.77\pm0.74$ & $+0.38\pm1.00$ & $56.70\pm9.43$\,ms & $0.925\pm0.123\times$ & $-5.95\%$ \\
        BoolQ, ordinary attn. & 64 & $71.40\pm0.22$ & --- & $67.36\pm0.51$\,ms & 1.000$\times$ & 0.00\% \\
        BoolQ, fixed 25\% & 64 & $70.62\pm0.55$ & $-0.77\pm0.56$ & $70.91\pm7.25$\,ms & $0.967\pm0.083\times$ & $-4.63\%$ \\
        BoolQ, fixed 50\% & 64 & $70.84\pm0.49$ & $-0.56\pm0.70$ & $69.35\pm7.61$\,ms & $0.992\pm0.093\times$ & $-7.19\%$ \\
        QuALITY 16K, ordinary attn. & 4 & $26.59\pm0.44$ & --- & $578.72\pm2.16$\,ms & 1.000$\times$ & 0.00\% \\
        QuALITY 16K, fixed 25\% & 4 & $26.09\pm0.40$ & $-0.50\pm0.63$ & $552.90\pm0.90$\,ms & $1.047\pm0.005\times$ & $-2.12\%$ \\
        QuALITY 16K, fixed 50\% & 4 & $26.94\pm0.62$ & $+0.35\pm0.18$ & $517.88\pm3.71$\,ms & $1.118\pm0.010\times$ & $-3.92\%$ \\
        \bottomrule
    \end{tabular}
\end{table*}

Table~\ref{tab:finetuning-schedules} reports results when replacement is applied during finetuning rather than inherited from pretraining.
One-time replacement incurs a modest one-off cost but does not improve accuracy or update time.
The train-loss-triggered schedule repeatedly measures attention and is especially costly on BoolQ.
Post-training replacement avoids training-time kernel overhead but provides no adaptation after replacement.

\begin{table*}[!htbp]
    \centering
    \small
    \caption{Replacing 25\% of heads under three RoPE finetuning schedules.}
    \label{tab:finetuning-schedules}
    \begin{tabular}{llrrr}
        \toprule
        Task & Schedule & $\Delta$ accuracy (pp) & Time/update & Controller cost \\
        \midrule
        SST-2 & One-time & $-0.19\pm0.40$ & 33.27\,ms & 0.70\,s \\
              & Train-loss & $-0.65\pm0.77$ & 36.15\,ms & 45.12\,s \\
              & Post-training & $-1.03\pm0.24$ & 30.75\,ms & 0.71\,s \\
        BoolQ & One-time & $-1.27\pm1.29$ & 40.85\,ms & 6.18\,s \\
              & Train-loss & $-0.38\pm0.15$ & 66.34\,ms & 70.15\,s \\
              & Post-training & $-1.03\pm0.33$ & 31.44\,ms & 6.17\,s \\
        \bottomrule
    \end{tabular}
\end{table*}

Zero-shot evaluations on HellaSwag \citep{zellers-etal-2019-hellaswag},
PIQA \citep{bisk2020piqa}, ARC-Easy \citep{clark2018think},
SST-2 \citep{socher-etal-2013-recursive}, and
BoolQ \citep{clark-etal-2019-boolq}
find no Holm-corrected significant change after 25\% replacement; this does not establish performance equivalence.

\subsection{1B training and evaluation}
\label{app:scale-up-details}

\paragraph{Pretraining and perplexity.}
The 32-layer decoder has 16 attention heads per layer, hidden dimension 1,536, an MLP expansion factor of four, tied token embeddings, and no biases or dropout.
It contains 983,336,448 trainable parameters; an unused frozen absolute-position table retained for checkpoint compatibility brings the stored parameter count to 995,919,360.
RoPE is used throughout, with base 10,000 and context length 8,192.
Four GH200 GPUs each use micro-batch size two and eight accumulation steps, giving 524,288 target tokens per optimiser update without activation checkpointing.
Training performs 37,512 updates, processing 19,667,091,456 target tokens from a shuffled non-overlapping FineWeb-Edu corpus without repeating target positions.
AdamW uses a peak learning rate of $3\times10^{-4}$, 375 linear warm-up updates, cosine decay to $3\times10^{-5}$, $(\beta_1,\beta_2)=(0.9,0.95)$, weight decay 0.1, and gradient clipping at norm 1.

All four configurations share pretraining seed 1337 and the update-18,756 midpoint checkpoint, including optimiser and sampler state.
Replacement selects 128 or 256 of the 512 heads by attention variance and fits their post-softmax means with the absolute-plus-relative representation.
Calibration samples 16 sequences of length 8,192 from the training corpus and measures the current checkpoint on rank zero without updating its weights or advancing the training sampler; fitting uses 400 steps and the same 0.2-nat mean row-wise KL acceptance threshold as the 124M experiments.
Table~\ref{tab:intervention-costs} reports calibration, fitting, and total intervention time.
The pruning control applies the gate-Taylor criterion in Equation~\ref{eq:gate-taylor-pruning} to the same 16 calibration sequences and removes the 128 lowest-scoring heads.
It physically removes the corresponding projection rows and columns, retaining the remaining AdamW state, learning-rate schedule, and data order through update 37,512.
Final perplexities in Table~\ref{tab:scale-up-results} use 512 sequential 8K windows, or 4,194,304 target tokens, from a reporting interval disjoint from calibration.
Pretraining results use one seed; the downstream evaluations use three finetuning seeds for each resulting checkpoint.

\paragraph{Post-replacement update measurements.}
Table~\ref{tab:scale-up-systems} distinguishes four-GPU update time, including gradient communication, from single-GPU time and memory.
The distributed benchmark uses the training configuration: four GH200 GPUs, $B=2$ per rank, eight accumulation steps, and 524,288 tokens per update.
Four pairs per replacement rate alternate execution order; each block reloads the source weights, starts fresh benchmark AdamW state, and performs ten warm-up and five measured updates.
Only one model is resident per block, and all models use identical preloaded native-length batches.
Update time is the slowest rank's wall time, including forward and backward computation, clipping, AdamW, and gradient all-reduce, but excluding loading, calibration, warm-up, and checkpoint I/O.
Reported times are means across blocks; speedups are means of within-pair ratios with SE across the four pairs, not pretraining seeds.

The single-GPU benchmark uses four fresh-process pairs, two in each execution order, with three warm-up and ten timed optimiser updates per model in each pair.
At $B=1$ and $B=2$, accumulation counts of 16 and eight keep the workload at 131,072 tokens per update, matching the per-rank pretraining workload.
Ordinary-attention heads use forced FlashAttention, and fused dispatch is verified for replaced heads.
Reported times are medians across pairs; speedups are medians of within-pair ratios, and peak-memory changes compare the paired configurations at the same batch size.
Separate ordinary-attention measurements accompany the two replacement rates.
These measurements include gradient accumulation and the optimiser update but exclude data loading, distributed communication, calibration, and pattern fitting.

\begin{table}[!htbp]
    \centering
    \footnotesize
    \setlength{\tabcolsep}{4pt}
    \caption{1B post-replacement update cost at 8K. Four-GPU speedup errors are SE over four timing pairs; single-GPU results use medians.}
    \label{tab:scale-up-systems}
    \begin{tabular}{@{}lcrrrr@{}}
        \toprule
        GPUs / $B$ & Rate & Ordinary attn. (ms) & \ourapproach{} (ms) & Speedup ($\times$) & Peak change (\%) \\
        \midrule
        \multicolumn{6}{@{}l}{\emph{Four GPUs, including communication: 524,288 tokens per update}} \\
        4 / 2 & 25\% & 4258.67 & 3986.67 & $1.0682\pm0.0005$ & --- \\
        4 / 2 & 50\% & 4263.01 & 3653.78 & $1.1667\pm0.0003$ & --- \\
        \midrule
        \multicolumn{6}{@{}l}{\emph{Single GPU, no communication: 131,072 tokens per update}} \\
        1 / 1 & 25\% & 4224.85 & 3958.26 & 1.067 & $-0.57$ \\
        1 / 1 & 50\% & 4226.87 & 3632.25 & 1.163 & $-2.31$ \\
        1 / 2 & 25\% & 4081.61 & 3828.21 & 1.066 & $-1.83$ \\
        1 / 2 & 50\% & 4083.50 & 3505.94 & 1.165 & $-3.99$ \\
        \bottomrule
    \end{tabular}
\end{table}

\paragraph{Downstream finetuning.}
Table~\ref{tab:scale-up-results} summarises quality for the four pretrained checkpoints; Table~\ref{tab:scale-up-downstream} gives individual finetuning seeds, paired changes, and observed update times.
Training uses causal-LM answer-token supervision with learning rate $2\times10^{-5}$, AdamW weight decay 0.01, gradient clipping at norm 1, and bfloat16.
For each task and seed, the ordinary-attention model selects an epoch from one through five on a stratified 10\% internal development split; all alternatives train for exactly that many epochs on the same examples in the same order before evaluation on the official validation set.
Selected epochs are 5/1/1 for SST-2, 4/2/4 for BoolQ, and 5/5/5 for QuALITY.
The pretrained head masks, patterns, and pruned layouts are inherited unchanged, with no additional selection or fitting.
SST-2, BoolQ, and QuALITY use token caps of 128, 512, and 8,192, and micro-batches of 128, 32, and four, respectively.
Only QuALITY accumulates gradients over four micro-batches; mean training-example lengths are 25.4, 142.8, and 6,207.5 tokens.

The paired accuracy differences between \ourapproach{} 25\% and gate-Taylor pruning are $-0.08\pm0.04$, $-0.19\pm2.07$, and $-0.06\pm0.75$ percentage points on SST-2, BoolQ, and QuALITY, respectively.
These three finetuning repetitions measure task-training variation, not equivalence between methods or variation across pretraining seeds.

We measure 1B finetuning time within the training loop up to the selected epoch, including batching and optimisation and excluding development evaluation.
Each seed contributes its mean time per optimiser update; the table reports the mean and standard error across seeds.
Runs use different GPU allocations: all pruning runs and most 25\% replacement runs reuse previously completed ordinary-attention references.
Pruning, 25\% replacement, and two QuALITY seed groups use \texttt{expandable\_segments:True}; earlier runs use the default allocator.
Successful replacements for two out-of-memory QuALITY attempts retain the same context, batch size, and task protocol.
The table therefore reports observed training times rather than controlled paired speedups; Table~\ref{tab:scale-up-systems} uses controlled pairs for pretraining-update costs.

\begin{table}[!htbp]
    \centering
    \footnotesize
    \setlength{\tabcolsep}{4pt}
    \caption{1B finetuning by seed: accuracy (\%), paired change from ordinary attention (pp), and observed mean update time.
    Errors are SE across finetuning seeds.}
    \label{tab:scale-up-downstream}
    \begin{tabular}{@{}llrrrrr@{}}
        \toprule
        & & \multicolumn{3}{c}{Accuracy by seed $\uparrow$} & & \\
        \cmidrule(lr){3-5}
        Task & Model & 1337 & 1338 & 1339 & Change (pp) & Update (ms) \\
        \midrule
        SST-2 & Ordinary attention & 91.63 & 90.94 & 91.74 & --- & $173.58\pm0.74$ \\
         & \ourapproach{} 25\% & 90.94 & 91.86 & 92.09 & \qimprove{$+0.19\pm0.47$} & $166.78\pm2.75$ \\
         & Pruning 25\% & 91.06 & 91.86 & 92.20 & \qimprove{$+0.27\pm0.44$} & $156.60\pm0.97$ \\
         & \ourapproach{} 50\% & 91.28 & 91.51 & 90.14 & \qdegrade{$-0.46\pm0.63$} & $158.36\pm3.67$ \\
        \midrule
        BoolQ & Ordinary attention & 75.99 & 76.39 & 75.57 & --- & $222.70\pm0.32$ \\
         & \ourapproach{} 25\% & 72.39 & 77.03 & 76.39 & \qdegrade{$-0.71\pm1.45$} & $215.11\pm0.39$ \\
         & Pruning 25\% & 75.69 & 73.30 & 77.40 & \qdegrade{$-0.52\pm1.43$} & $200.03\pm0.74$ \\
         & \ourapproach{} 50\% & 75.17 & 76.24 & 75.63 & \qdegrade{$-0.31\pm0.27$} & $196.13\pm0.59$ \\
        \midrule
        QuALITY & Ordinary attention & 24.98 & 28.33 & 29.00 & --- & $3504.01\pm5.63$ \\
         & \ourapproach{} 25\% & 27.71 & 27.90 & 27.18 & \qimprove{$+0.16\pm1.35$} & $3267.23\pm10.13$ \\
         & Pruning 25\% & 26.27 & 28.62 & 28.09 & \qimprove{$+0.22\pm0.64$} & $2972.08\pm1.10$ \\
         & \ourapproach{} 50\% & 26.32 & 23.73 & 25.31 & \qdegrade{$-2.32\pm1.85$} & $2989.63\pm5.40$ \\
        \bottomrule
    \end{tabular}
\end{table}

\FloatBarrier
\subsection{Replacement in pretrained models}
\label{app:pretrained-static-details}

Qwen3-4B has 36 layers, 32 query heads, eight key/value heads, and head dimension 128.

\subsubsection{Qwen3-4B zero-shot task accuracy}
We evaluate the five zero-shot tasks using loglikelihood over the full validation split of each task, comparing candidate continuations without generation and grouping requests by exact token length so that no padding enters the fused kernel.
Both selectors use the same fitting protocol and 512 FineWeb-Edu calibration sequences, with compact patterns fitted separately for their selected head sets.
The median row-wise fit KL values are similar, at 0.027 nats for variance-selected heads and 0.029 nats for forward-KL-selected heads.

\begin{table}[!htbp]
    \centering
    \small
    \setlength{\tabcolsep}{4.5pt}
    \caption{Qwen3-4B zero-shot accuracy (\%) with compact FineWeb-Edu mean patterns.
    The last row gives the mean change from the ordinary-attention model in percentage points.}
    \label{tab:qwen-zeroshot}
    \begin{tabular}{lrrrrrrr}
        \toprule
        & & \multicolumn{3}{c}{Attention variance} & \multicolumn{3}{c}{Forward KL} \\
        \cmidrule(lr){3-5}\cmidrule(l){6-8}
        Task & Ordinary attn. & 10\% & 20\% & 30\% & 10\% & 20\% & 30\% \\
        \midrule
        HellaSwag & 52.28 & 52.02 & 51.21 & 48.92 & 49.10 & 45.05 & 43.81 \\
        PIQA & 74.86 & 75.19 & 74.48 & 73.56 & 74.70 & 72.58 & 70.40 \\
        ARC-Easy & 80.43 & 79.63 & 77.61 & 75.72 & 72.90 & 68.06 & 64.10 \\
        SST-2 & 88.65 & 88.65 & 87.96 & 85.44 & 87.61 & 63.07 & 68.81 \\
        BoolQ & 84.46 & 83.33 & 80.49 & 73.12 & 80.03 & 71.38 & 71.38 \\
        \midrule
        Mean change & --- & $-0.37$ & $-1.79$ & $-4.79$ & $-3.27$ & $-12.11$ & $-12.44$ \\
        \bottomrule
    \end{tabular}
\end{table}

Variance selection gives higher accuracy in all fifteen task--rate comparisons in Table~\ref{tab:qwen-zeroshot}.
The selected sets overlap in 31\% of heads at 10\% replacement and 43\% at 30\%.
At 10\%, the counts over layers 1--12, 13--24, and 25--36 are 50/18/47 for forward KL and 37/31/47 for variance.
The variance-selected heads also have a higher median per-input forward-KL score, 0.578 versus 0.296 nats at 10\% replacement, despite giving higher task accuracy when replaced together.
These observations show that the scores select different head sets; neither the layer counts nor the pattern distances establish which head functions cause the accuracy differences.

SST-2 accuracy under forward-KL selection falls by approximately 25.6 percentage points at 20\% replacement, then recovers partially at 30\%.
The prediction counts for its two labels change from 479/393 in the ordinary-attention model to 721/151 at 20\% and 580/292 at 30\%, compared with gold counts of 428/444.
Thus, the largest accuracy loss coincides with the strongest imbalance towards one predicted label, but these aggregate counts do not establish the mechanism behind the non-monotone curve.
Under variance selection, the SST-2 accuracy loss is at most 3.21 points across the tested rates.

\FloatBarrier
\subsubsection{GQA prefill measurements}

The prefill benchmark compares the fused GQA path with Flash scaled dot-product attention at $B\in\{4,8\}$ and $T\in\{2048,4096,8192\}$.
It uses forward-KL-selected heads and patterns fitted on 128 FineWeb-Edu sequences of 2,048 tokens.
Table~\ref{tab:qwen-prefill-details} reports end-to-end and attention-operation speedups for these systems configurations, separately from the 512-sequence task evaluation above.
Measurements beyond the calibration length assess execution speed.

\begin{table*}[!htbp]
    \centering
    \small
    \caption{Qwen3-4B prefill speedup ($\times$): end-to-end, with attention-operation speedup in parentheses.}
    \label{tab:qwen-prefill-details}
    \begin{tabular}{rrrrrrr}
        \toprule
        $B$ & $T$ & 10\% & 20\% & 30\% & 50\% & 75\% \\
        \midrule
        4 & 2048 & 1.011 (1.16) & 1.028 (1.18) & 1.068 (1.21) & 1.135 (1.29) & 1.225 (1.38) \\
        4 & 4096 & 1.027 (1.16) & 1.057 (1.19) & 1.097 (1.23) & 1.156 (1.28) & 1.246 (1.38) \\
        4 & 8192 & 1.042 (1.15) & 1.072 (1.19) & 1.115 (1.23) & 1.184 (1.30) & 1.250 (1.37) \\
        8 & 2048 & 1.015 (1.15) & 1.042 (1.17) & 1.079 (1.21) & 1.148 (1.28) & 1.231 (1.37) \\
        8 & 4096 & 1.029 (1.15) & 1.061 (1.19) & 1.080 (1.18) & 1.156 (1.27) & 1.252 (1.37) \\
        8 & 8192 & 1.040 (1.15) & 1.071 (1.18) & 1.110 (1.22) & 1.178 (1.30) & 1.272 (1.40) \\
        \bottomrule
    \end{tabular}
\end{table*}

\section{Kernel Implementation and Validation}
\label{app:kernel-details}

\FloatBarrier
\subsection{Compact fitting and storage}

The need for compact storage is particularly clear at long context lengths: at $T=10^{6}$, one causal fp32 matrix occupies approximately 2\,TB even if only its lower triangle is stored.
Equation~\ref{eq:fit-objective} is the maximum-likelihood objective of a log-linear model and is convex in $(\alpha,\rho)$ up to additive shifts absorbed by row normalisation.
At an optimum, the fitted prior reproduces both target marginals because
\begin{equation}
    \frac{\partial\mathcal{L}_{\mathrm{fit}}}{\partial\alpha(j)}
    =\frac{1}{T}\left(\sum_i\widehat{\prior}(i,j)-\eta_{\mathrm{abs}}(j)\right),
\end{equation}
with the analogous condition for $\rho$.
The fitting settings and acceptance threshold are given in Appendix~\ref{app:controlled-pretraining-details}.
The dense implementation explicitly forms causal rows; the FFT implementation evaluates the convolutional row normalisers from the sufficient statistics. These fitting paths are distinct from reconstruction inside the fused execution kernel.
For the 4K RoPE head set in Table~\ref{tab:implementation-validation}, the dense implementation allocates an fp32 $H\times T\times T$ buffer in each of the ten affected layers, including unused head slots, for 8.060\,GB in total.
The compact buffers occupy 0.133\,GB; an ideally packed dense representation containing only the 36 selected matrices would occupy 2.416\,GB.

\FloatBarrier
\subsection{Mixed-head execution and numerical checks}
\label{sec:implementation-validation}

The fused kernel executes ordinary-attention and replaced heads in a single forward launch per layer, writing outputs in their original head order.
Each kernel program handles one batch element, head, and block of query positions; a head-state flag selects the computation for that program.
Ordinary-attention heads use the online-softmax tiling of FlashAttention-2 \citep{dao2022flashattention,dao2023flashattention2}.
For a replaced head, the program reconstructs each causal tile of $\widehat{\prior}$ in registers from $\alpha$, $\rho$, and precomputed row normalisers, multiplies it with the value tile, and accumulates the output.
The fixed path neither computes query-key scores nor reads a dense $T\times T$ pattern, and preserving head order avoids concatenating separate ordinary-attention and replaced-head outputs.

For fixed attention weights, the attention operation's backward pass is $\mathrm{d}\mathbf{V}=\widehat{\prior}^{\top}\mathrm{d}\mathbf{Z}$.
Key-block programs accumulate value gradients for every head and key gradients for ordinary-attention heads; a separate query-block pass computes query gradients only for ordinary-attention heads.
Our multi-head pretraining implementation computes query and key projections only for ordinary-attention heads.
The Qwen grouped-query implementation omits replaced query projections but retains the full shared key and value projections.

Persistent state per replaced head consists of $\alpha$, $\rho$, the normalisers, and a small aligned relative-distance band for coalesced tile loads, using $O(T)$ storage.
The multiplication $\widehat{\prior}\mathbf{V}$ still costs $O(T^2d_h)$: savings come from removing score and softmax computation and reducing pattern bandwidth and activation storage.

Table~\ref{tab:implementation-validation} combines pattern fidelity, storage, and forward checks; Table~\ref{tab:gradient-errors} reports gradient errors against the eager reference.

\begin{table*}[!htbp]
    \centering
    \small
    \caption{Representation quality at 25\% replacement (one seed), prior storage, and kernel checks.}
    \label{tab:implementation-validation}
    \label{tab:representation-details}
    \begin{tabular}{@{}lrrrr@{}}
        \toprule
        Position & Ordinary attn. PPL & Dense PPL & Abs.+rel. PPL & Storage reduction \\
        \midrule
        Absolute & 26.0252 & 26.1184 & 26.1150 & 59.9$\times$ \\
        RoPE & 23.9571 & 24.1180 & 24.1165 & 60.6$\times$ \\
        \bottomrule
    \end{tabular}
    \par\medskip
    \begin{tabular}{llll}
        \toprule
        Component & Configuration & Metric & Result \\
        \midrule
        Prior buffers & RoPE, 36 heads in 10 layers & Dense impl. / abs.+rel. & 8.060 / 0.133\,GB \\
        Mixed-head forward & nanoGPT, FP32 & Maximum absolute error & $1.07\times10^{-6}$ \\
        State transitions & 50-operation fuzz test & Worst oracle error & $9.54\times10^{-7}$ \\
        Mixed-head backward & Replaced / ordinary-attention Q,K rows & Gradient status & zero / nonzero \\
        Fused dispatch & nanoGPT, BF16 & Triton calls & 2 / 2 \\
        \bottomrule
    \end{tabular}
\end{table*}

We check the fused implementation against an eager reference in fp32 and bfloat16.
The suite covers empty, partially replaced, and fully replaced head sets; unaligned sequence lengths; non-contiguous query/key views; head dimensions 64, 96, and 128; and mixtures of dense and absolute-plus-relative patterns.
Replaced heads have no query/key slot on the multi-head long-context path and therefore receive zero query/key gradients by construction.
The grouped-query path maps each ordinary-attention query head to its retained shared key/value group.
Table~\ref{tab:gradient-errors} reports numerical errors for a mixed-head RoPE layer on GH200, comparing fused execution with an eager dense reference that reconstructs the same stored compact patterns.
The test uses batch size two, sequence length 129, four heads of dimension 64, and a stored prior length of 192; heads 1 and 3 are fixed, using zero-based indices.
Both paths receive the same output gradient, and fused dispatch is checked.
Parameters remain FP32; the BF16 test uses autocast, as in training.
For reference tensor $u$ and fused result $\widetilde{u}$, we report maximum absolute error and relative $L_2$ error, $\lVert\widetilde{u}-u\rVert_2/\max(\lVert u\rVert_2,10^{-8})$.
The query/key weight comparisons include the ordinary-attention heads' gradients and the zero rows of replaced heads; value gradients include both head types.
The same seven-component test for pruning gives maximum relative $L_2$ errors of $7.48\times10^{-7}$ in FP32 and $4.45\times10^{-3}$ under BF16 autocast.

\begin{table}[!htbp]
    \centering
    \small
    \caption{Numerical agreement with the eager reference for a layer with ordinary-attention and fixed-mean heads.
    Projection rows report weight gradients.}
    \label{tab:gradient-errors}
    \begin{tabular}{@{}lrrrr@{}}
        \toprule
        & \multicolumn{2}{c}{FP32} & \multicolumn{2}{c}{BF16 autocast} \\
        \cmidrule(lr){2-3}\cmidrule(l){4-5}
        Tensor & Max. absolute & Relative $L_2$ & Max. absolute & Relative $L_2$ \\
        \midrule
        Output & $1.19\times10^{-7}$ & $1.86\times10^{-7}$ & $3.91\times10^{-3}$ & $2.80\times10^{-3}$ \\
        Input gradient $dX$ & $3.58\times10^{-7}$ & $2.51\times10^{-7}$ & $7.81\times10^{-3}$ & $3.68\times10^{-3}$ \\
        Value gradient $dV$ & $3.58\times10^{-7}$ & $1.78\times10^{-7}$ & $7.81\times10^{-3}$ & $1.46\times10^{-3}$ \\
        Query projection & $8.34\times10^{-7}$ & $2.82\times10^{-7}$ & $1.56\times10^{-2}$ & $4.53\times10^{-3}$ \\
        Key projection & $1.19\times10^{-6}$ & $3.01\times10^{-7}$ & $1.56\times10^{-2}$ & $4.43\times10^{-3}$ \\
        Value projection & $3.81\times10^{-6}$ & $2.30\times10^{-7}$ & $6.25\times10^{-2}$ & $2.05\times10^{-3}$ \\
        Output projection & $3.81\times10^{-6}$ & $1.97\times10^{-7}$ & $6.25\times10^{-2}$ & $2.68\times10^{-3}$ \\
        \bottomrule
    \end{tabular}
\end{table}

\FloatBarrier
\subsection{Causal-prefill benchmarks}

The checkpoint-backed prefill benchmark compares the seed-1337 ordinary-attention 16K RoPE checkpoint under forced FlashAttention with its 25\%- and 50\%-replaced counterparts.
Inputs are seeded token tensors, and each timing point is the median of four or six fresh-process ABBA pairs without CUDA-Graph replay.
The input-length sweep fixes $B=64$ over $T\in\{512,1024,2048,4096,8192,16384\}$, while the batch-size sweep fixes $T=4096$ over $B\in\{2,4,8,16,32,64\}$.
A separately collected launch-bound $B=1$ measurement is retained only as a diagnostic.

\begin{table}[!htbp]
    \centering
    \small
    \caption{Selected causal-prefill latencies for the ordinary-attention and replaced 16K RoPE checkpoints at $B=64$, without CUDA Graph replay.
    Input length varies while the checkpoints remain fixed.}
    \label{tab:prefill-latency}
    \begin{tabular}{@{}lccrrr@{}}
        \toprule
        Input length & Batch size $B$ & Replaced heads & Ordinary attn. (ms) & Replaced (ms) & Speedup \\
        \midrule
        4K & 64 & 25\% & 306.70 & 278.33 & 1.100$\times$ \\
        4K & 64 & 50\% & 303.94 & 247.47 & 1.229$\times$ \\
        8K & 64 & 25\% & 776.17 & 705.03 & 1.100$\times$ \\
        8K & 64 & 50\% & 760.20 & 622.94 & 1.221$\times$ \\
        16K & 64 & 25\% & 2075.04 & 1900.13 & 1.092$\times$ \\
        16K & 64 & 50\% & 2045.38 & 1703.87 & 1.200$\times$ \\
        \bottomrule
    \end{tabular}
\end{table}

Figure~\ref{fig:systems-across-stages}c,d shows the checkpoint-backed causal-prefill speedups across input lengths and batch sizes; Table~\ref{tab:prefill-latency} gives selected absolute latencies.
Latency columns are medians across pairs, whereas speedups are medians of within-pair ratios and need not equal the ratios of the displayed latency medians.
The figure's whiskers span the observed pairs.
At fixed batch size 64, the checkpoint-backed benchmark remains above break-even over the complete 512--16K input-length range.
The measured ranges are 1.092--1.116$\times$ at 25\% replacement and 1.200--1.244$\times$ at 50\%.
At fixed 4K length, the primary batch sweep uses sizes two through 64 and is similarly stable.

The separately retained batch-one point is dominated by launch overhead and exhibits large process-to-process variation. At $T=4096$ and $B=1$, the median paired speedups are 0.969$\times$ and 0.972$\times$ at 25\% and 50\% replacement, respectively; their observed ranges are 0.878--1.067$\times$ and 0.599--1.498$\times$ across six pairs.

We retain measurements only when the zero-replacement control remains within the contention tolerance of 1.00$\times$.

In batch-one shape benchmarks, the 124M shape at 16K reaches 1.08$\times$ causal-prefill speedup with 25\% replacement and 1.33$\times$ with full replacement.
For a 774M shape, the 25\% speedup is 1.07$\times$ at 8K and 1.08$\times$ at 16K, while full replacement reaches 1.31$\times$ and 1.36$\times$.
CUDA-Graph replay isolates launch overhead for the checkpoint-backed 4K model: at batch size one, 25\%, 50\%, and 75\% replacement reach 1.113$\times$, 1.220$\times$, and 1.377$\times$ speedup, respectively.
These are forward-only kernel measurements.
They do not measure model perplexity, token-by-token decoding, K-cache removal, or static query/key parameter repacking.

\section{Data, Models, and Licences}
\label{app:resources}

Table~\ref{tab:resource-licences} lists the external datasets and pretrained weights used in the reported experiments, including the auxiliary zero-shot evaluations in Appendix~\ref{app:downstream-details}.
Asset names link to the repositories from which they can be accessed; licence entries link to the corresponding upstream statements.
We train the 124M models in the controlled pretraining study from scratch.

\begin{table}[ht]
    \centering
    \small
    \setlength{\tabcolsep}{5pt}
    \renewcommand{\arraystretch}{1.15}
    \caption{External datasets and pretrained models.}
    \label{tab:resource-licences}
    \begin{tabular}{@{}p{0.27\linewidth}p{0.25\linewidth}p{0.40\linewidth}@{}}
        \toprule
        Asset & Use & Upstream licence or terms \\
        \midrule
        \multicolumn{3}{@{}l}{\textit{Datasets}} \\
        \href{https://huggingface.co/datasets/HuggingFaceFW/fineweb-edu}{FineWeb-Edu} & Pretraining; calibration and evaluation & \href{https://huggingface.co/datasets/HuggingFaceFW/fineweb-edu\#licensing-information}{ODC-By 1.0}; Common Crawl terms \\
        \href{https://huggingface.co/datasets/nyu-mll/glue}{SST-2 (GLUE)} & Finetuning and zero-shot evaluation & \href{https://huggingface.co/datasets/stanfordnlp/sst2\#licensing-information}{Not specified in the original dataset card} \\
        \href{https://huggingface.co/datasets/aps/super_glue}{BoolQ (SuperGLUE)} & Finetuning and zero-shot evaluation & \href{https://github.com/google-research-datasets/boolean-questions\#license}{CC BY-SA 3.0} \\
        \href{https://huggingface.co/datasets/tasksource/QuALITY}{QuALITY} & Long-input finetuning & \href{https://nyu-mll.github.io/quality/}{CC BY 4.0}; article-level licences \\
        \href{https://huggingface.co/datasets/Rowan/hellaswag}{HellaSwag} & Zero-shot evaluation & \href{https://github.com/rowanz/hellaswag/blob/master/LICENSE}{MIT} \\
        \href{https://huggingface.co/datasets/ybisk/piqa}{PIQA} & Zero-shot evaluation & \href{https://github.com/ybisk/ybisk.github.io/blob/master/piqa/README.md}{Academic Free License 3.0} \\
        \href{https://huggingface.co/datasets/allenai/ai2_arc}{ARC-Easy} & Zero-shot evaluation & \href{https://huggingface.co/datasets/allenai/ai2_arc/blob/main/README.md}{CC BY-SA 4.0} \\
        \midrule
        \multicolumn{3}{@{}l}{\textit{Pretrained weights}} \\
        \href{https://huggingface.co/Qwen/Qwen3-4B}{Qwen3-4B} & Static replacement and prefill & \href{https://huggingface.co/Qwen/Qwen3-4B}{Apache 2.0} \\
        \bottomrule
    \end{tabular}
\end{table}

The FineWeb-Edu database licence does not replace the rights in the underlying web content; the release also refers users to \href{https://commoncrawl.org/terms-of-use/}{Common Crawl's terms of use}.
QuALITY records a separate licence for each source article in its \href{https://github.com/nyu-mll/quality}{upstream data}.
GLUE directs users to the original dataset licences; its loading code does not establish a licence for SST-2's review text.

The implementation builds on \href{https://github.com/karpathy/nanoGPT/blob/master/LICENSE}{nanoGPT} and \href{https://github.com/triton-lang/triton/blob/main/LICENSE}{Triton}, both under MIT licences.
The 124M experiments use the GPT-2 tokeniser through \href{https://github.com/openai/tiktoken/blob/main/LICENSE}{tiktoken} (MIT); the Qwen experiments use the tokeniser distributed with Qwen3-4B.
The synthetic MQAR data use the \href{https://github.com/HazyResearch/zoology/tree/1ad20d193b6113cae1e8f3c655c300d7b4b3f4bb}{Zoology generator}, distributed under \href{https://github.com/HazyResearch/zoology/blob/1ad20d193b6113cae1e8f3c655c300d7b4b3f4bb/LICENSE.md}{Apache 2.0}.

\end{document}